# Trends and Challenges in Authorship Analysis: A Review of ML, DL, and LLM Approaches

Nudrat Habib, Tosin Adewumi, Marcus Liwicki & Elisa Barney
ML Group, EISLAB, Luleå University of Technology, Sweden.
{firstname.lastname}@ltu.se

Authorship analysis plays an important role in diverse domains, including forensic linguistics, academia, cybersecurity, and digital content authentication. This paper presents a systematic literature review on two key sub-tasks of authorship analysis; Author Attribution and Author Verification. The review explores SOTA methodolo- gies, ranging from traditional ML approaches to DL models and LLMs, highlighting their evolution, strengths, and limitations, based on studies conducted from 2015 to 2024. Key contributions include a comprehensive analysis of methods, techniques, their corresponding feature extraction techniques, datasets used, and emerging chal- lenges in authorship analysis. The study highlights critical research gaps, particularly in low-resource language processing, multilingual adaptation, cross-domain generaliza- tion, and AI-generated text detection. This review aims to help researchers by giving an overview of the latest trends and challenges in authorship analysis. It also points out possible areas for future study. The goal is to support the development of better, more reliable, and accurate authorship analysis system in diverse textual domain.

## 1 Introduction

Authorship analysis involves studying the unique characteristics of a text to determine its author. This process is rooted in the field of linguistic research known as stylom- etry (El & Kassou, 2014). Research in authorship analysis dates back to 1966 when Bayesian statistical methods were employed by Watson (1966) to study the author- ship of the Federalist Papers, a collection of essays from the late 18th century. Since 1966, it has evolved significantly, transitioning from traditional statistical methods to advanced computational techniques. The study of authorship analysis is a broad dis- cipline with many sub-fields, each with its own applications, methods, and areas of interest. The key sub-fields of authorship analysis include Author Attribution (AA), Author Verification (AV), Authorship Discrimination (AD), cross-domain Authorship Analysis, Author Profiling and Author Obfuscation among others. All these tasks are important in various contexts like literary analysis, forensic linguistics, plagiarism detection, and digital content moderation (Peng et al., 2016). This literature review mainly looks at AA and AV. AA involves finding out who likely wrote a text by studying the language and writing style used (Rosen-Zvi et al., 2010) whereas AV checks if two texts were written by the same person (Halvani et al., 2016). There are two further sub-tasks of AV: human vs human AV and human vs machine AV.

AA and AV are essential tasks with many real-world applications in a variety of fields. For example, in the realm of cybersecurity, determining the author of dangerous or anonymous online information can help with fraud prevention, cybercrime detection, and online platform integrity (Jalal Yousef Zaidieh, 2024). In forensic linguistics, these techniques are crucial for solving legal cases, such as determining the authorship of anonymous letters, or verifying the authenticity of documents in criminal investigations (Ellen et al., 2018). Additionally, AA plays a significant role in digital content management by accurately assigning credit to creators, supporting copyright enforcement, and effectively detecting and preventing plagiarism in both academic and creative contexts (Lastowka, 2007). In literary studies, AA and AV contribute to understanding the historical and stylistic context of written works. By using linguistic and stylistic characteristics, researchers can assign disputed or anonymous works to certain writers, providing insight into intellectual history (B. Huang et al., 2024a).

Conducting a systematic literature review on AA and AV is crucial for synthesizing existing research, identifying gaps, and guiding future investigations. Although previous reviews have explored various aspects of this field, some areas remain under-explored or inadequately addressed. The rise of AI-generated content has introduced new challenges in distinguishing between human and machine authorship. Although existing reviews have addressed traditional AA, they often overlook the complexities introduced by AI-generated texts. One survey paper that focuses on both sub-tasks, is done by Tyo et al. (2022). This survey provides a comprehensive list of research in these sub-tasks, involving feature-based and embedding-based methods. Research involving LLMs is not part of that survey. Moreover, the most recent AV sub-task involving human vs machine-generated text is also not found. Another survey focusing on AA was done by He et al. (2024), where the current techniques and research gaps were discussed. However, it focuses on only one sub-task, and research leveraging LLM for the task is also not found in the survey. Our survey fills these gaps by providing most recent research involving recent advancements, technologies used and issues arising i.e. machine generated text. In this paper, we reviewed different methods used for AA and AV, their key classifications, feature extraction methods and their impact on classification, available datasets, challenges and limitations based on studies conducted from 2015 to 2024. This detailed analysis is a helpful guide for beginners who want to learn about and understand the fields of AA and AV.

## 1.1 Research questions and objective

This review aims to study the latest research in authorship analysis, with a focus on AA and AV using traditional ML, DL, and LLM. It categorizes studies by sub-tasks, summarizes datasets and methodologies, identifies challenges, and highlights research gaps to guide future investigations. The Research Questions (RQs) that we tried to answer are as follows:

- RQ1: What is the state-of-the-art(SOTA) in AA and AV?
- RQ2: Which ML , DL , and LLM-based techniques have been used for this purpose, based on studies conducted from 2015 to 2024.?
- RQ3: What are the current challenges in AA and AV?

## 1.2 Contributions

This paper makes the following key contributions:

1. A comprehensive systematic literature review of the recent SOTA in AA and AV.
2. Findings that provide insight from over 90 studies, grouped by sub-tasks of authorship analysis namely AA and AV including both human vs human and human vs machine verification.
3. A discussion on existing challenges, gaps, and future research directions is provided to help researchers identify areas for future research.

This paper is structured as follows; **Section 2** outlines the search methodology used and **Section 3** presents extensive literature analysis of all the included articles. **Section 4** discusses the key findings from this review, **Section 5** discusses potential research gaps and suggests future research directions in textual AA and AV and finally **Section 6** concludes this survey.

## 2 Search Methodology

The literature search was conducted manually using two primary databases: Google Scholar[1] and Scopus[2] . The search results were all properly documented See Appendix B, so that the review is replicable. The process is explained below.

1. Defined primary search terms based on the topic of literature review.
2. Developed specific search queries using the primary terms.
3. Accessed selected databases using these search queries to retrieve relevant articles.
4. Initially skimmed titles and abstracts to assess article relevance.
5. The initial analysis showed the queries were too broad, resulting in irrelevant results.
6. Revised and refined the primary search terms and queries to generate targeted results.
7. Repeated the search and refinement process three times until obtaining relevant, focused results.
8. After the third iteration, compiled a shortlist of approximately 150 research papers for further analysis.
9. Conducted detailed reading and analysis of each shortlisted article to determine inclusion in the systematic literature review.
10. During this step, research papers were identified which do not fall directly under the topic of the literature review and were excluded.

---

[1]https://scholar.google.com (last visited: October 27, 2024)
[2]https://www.scopus.com (last visited: October 27, 2024)

11. At the end of the this, we had a total of 93 research articles which were included in current systematic literature review (see Appendix B).

**Inclusion and exclusion criteria**:

1. Peer-reviewed articles on the topics of AA and AV using ML, DL, and LLM were included in the study.
2. Articles published in the last decade were included in the review, In addition, it will not be practical to make the timeline open-ended so a 10 year period was fixed, similar to Ouni et al. (2023) where they conducted survey for 12 years.
3. Articles written in English but analyzing text in other languages were included.
4. For AV, both human and GenAI verification is included in this review.
5. Due to the large number of documents, only the papers through the first 10 pages of the search were included in the study (after that articles also become less relevant).
6. The search resulted in all types of articles where AA was applied to different fields like chemistry, coding, or different modalities e.g. voice data, etc. All of these articles except the ones involving textual data were excluded.
7. Research articles published in any language other than English were also excluded.
8. The focus of our study is AA and AV, so research on other sub-tasks was excluded.
9. Due to the rise of Gen AI, AA for GenAI was also the topic of a few articles. Since the focus of this review is only AA involving humans, AA of GenAI is not included in this survey, except for the case where machine generated text was added as a single class to a dataset.

Eventually, after three iterations, 93 articles were selected. These articles were mostly peer-reviewed articles with a few preprint versions. Preprint versions were only added if they are most recent (from 2024) and secondly if they cover the most recent sub-task in AV i.e. human vs machine AV. The summary of venues is given in table 1.

## 3 Literature Analysis of selected papers in this survey

This section presents the techniques identified from the literature, that are used for the experiments. These techniques are divided into traditional ML-based, DL-based, LLM-based techniques and unsupervised methods and the detailed description of these techniques is provided in Appendix A. This section discusses in detail the current research using these techniques and their key features grouped by sub-tasks. Figure 1(a) illustrates the number and distribution of articles related to the AA sub-task, totaling 59 articles. Figure 1(b) presents the articles related to the AV sub-task, totaling 22 articles. Lastly, figure 1(c) displays the articles focusing on both sub-tasks, amounting to 12 articles.

**Table 1:** Venue Summary of All Articles

| Venue | Articles | Venue count |
|---|---|---|
| Preprint | 8 | 2 |
| Conference | 38 | 27 |
| Workshop | 2 | 2 |
| Journal | 45 | 30 |
| **Total** | **93** | **61** |

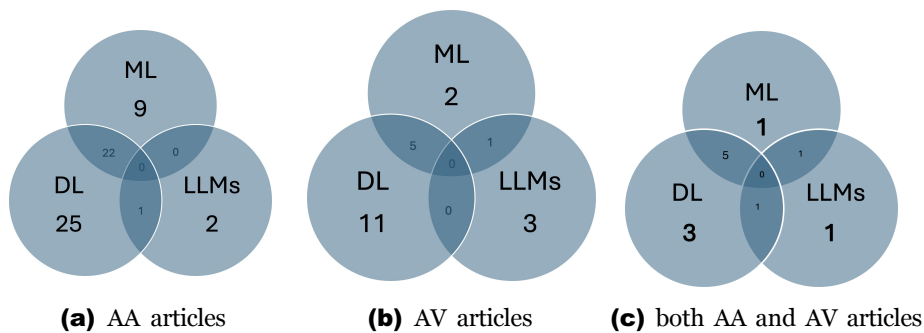

(a) AA articles  (b) AV articles  (c) both AA and AV articles

**Fig. 1**: Venn diagram showing number of articles and techniques grouped by sub-task

### 3.1 Author Attribution (AA)

AA is the most important sub-task of authorship analysis (Saedi and Dras, 2021). 59 out of 93 articles focused solely on this sub-task as shown in Figure 1a. Due to the large number of articles for this sub-task, we have divided the literature further based on the techniques used and will discuss them separately.

1. **Traditional ML-based**: Many Traditional ML algorithms have been employed for the task of AA. A total of 9 articles explored the effect and performance of these models for the task with Support Vector Machine (SVM) being the most prominent model and has been used by Marchenko et al. (2017), Khan et al. (2024), Altakrori et al. (2018), da Rocha et al. (2022), Raafat et al. (2021) and Fourkioti et al. (2019). Most of the studies involve multiple models at the same time and compare the performance for specific text type, domain or language. Zhang et al. (2018) explored an interesting use case of AA in historic documents for historically disputed documents using distributional semantics and suggested that Caesar probably did not write the African War, Alexandrine War, and Spanish War.

   Dimensionality reduction is important for improving model efficiency, reducing overfitting, and focusing on relevant features. It is done automatically in DL models but for classical ML models there are different techniques used. The study by Ríos-Toledo et al. (2023) dealt with the issue of high dimensionality using Principal Component Analysis (PCA) and Latent Semantic Analysis (LSA) as it decreases the performance of the classifier. The results showed that the reduction of dimensions with PCA and the use of Logistic Regression (LR) and SVM classifiers achieved better results than other similar works of the state of the art using the same corpus.

   An interesting approach to the problem is to combine the classifiers or features. Khan et al. (2024) ensembled SVM with boosted algorithms i.e. Gradient Boosting (GBC), AdaBoosting (ABC), CatBoosting (CBC), HistGradientBoosting (HGBC), LightGradientBoosting (LGBC) and XGBoosting (XGB) classification models, to create SVM-GBC, SVM-ABC, SVM-CBC, SVM-HGBC, SVM-LGBC and SVM-XGB respectively for AA in Urdu. The SVM-CBC model is seen as the most efficient one, with a success rate of 92%. Fourkioti et al. (2019) carried out a series of experiments with different classifier and feature based language models(LMs) and found that character or part of speech (POS) level language models achieve better classification results as compared to word level LM (as in literature). They also looked at combining features in learning algorithms by merging them, and at using language models built on different units by mixing their individual perplexity scores with a linear formula. Both approaches demonstrated effectiveness, achieving improved results compared to the individual feature classes or language models they combine. Using multiple features with an SVM classifier was done by Raafat et al. (2021) and Marchenko et al. (2017). Raafat et al. (2021) reported that adding syntactic features enhances the F1 score; in their experiments, they found that the f1 score averages per author increased by 0.3% when adding features. A comparison of instance based approach (stylometric features are extracted from each tweet, then the model is trained) and profile based approach (tweets from same users are used to create a file, then the model is trained) was done by Altakrori et al. (2018). Profile-based approaches were in line with the current SOTA but they argue that they are simpler and their results can be visualized in a more intuitive way in comparison to instant based approaches, which gives them an edge to be used in court (a use case where explainability is important). The summary of these articles is given in Table 2.

**Table 2:** Summary of Research on AA using Traditional ML

| # | Article | Model | Dataset | Language |
|---|---------|-------|---------|----------|
| 1 | Marchenko et al. (2017) | SVM | RCV1 | English |
| 2 | Zhang et al. (2018) | Random indexing, cosine similarity | War memoirs | Latin |
| 3 | Ríos-Toledo et al. (2023) | PCA, LSA, SVM, KNN, LR | PAN 2012 | English |
| 4 | Khan et al. (2024) | Variants of SVM (XGB, ABC, CBC, GBC, LGBC, HGBC) | UACV-22 | Urdu |
| 5 | Altakrori et al. (2018) | NB, SVM, DT, RF | New dataset | Arabic |
| 6 | Sarwar et al. (2024) | KNN, DT, NB, LR, DL, RF | New dataset | Sinhala |
| 7 | da Rocha et al. (2022) | SVM, LR | New dataset | Portuguese |
| 8 | Raafat et al. (2021) | SVM, NB, LR | Gungor 50-Author | English |
| 9 | Fourkioti et al. (2019) | SVM, KNN, RF, multinomial NB, n-gram LM (words, POS) | IMDB62, Tweet dataset | English |

2. **DL Based Research**: As presented in Table 3, CNN and BERT (including variants) are the predominant DL models for AA, with each model used in 10 articles each. The table summarizes the usage across a total of 25 articles, including these and additional deep learning models. Comparison of different models, feature extraction techniques, concatenation of these techniques and comparison of different embedding techniques were done by different researchers in context of AA and they reported their results.

**Table 3:** Summary of Research Articles on AA using DL

| # | Article | Model | Dataset | Language |
|---|---------|-------|---------|----------|
| 1 | Ribeiro et al. (2024) | Fuzzy fingerprint, RoBERTa | IMDb, Blog authorship | English |
| 2 | Silva et al. (2024) | Gan-BERT | Gutenberg novels | English |
| 3 | Rahgouy et al. (2024) | BERT | IMDB62, CCAT50, Blog50/1000, Arxiv100 | English |
| 4 | Silva and Frommholz (2023) | BERT, Multimodal transformer | ArXiv | English |
| 5 | Saputra and Riccosan (2024) | IndoBERT | News articles | Indonesian |
| 6 | Suman et al. (2021) | Capsule CNN, KNNs | Twitter | English |
| 7 | AlZahrani and Al-Yahya (2023) | AraBERT, AraELECTRA, ARBERT, MARBERT | New dataset | Arabic |
| 8 | Al-Sarem et al. (2023) | CNN, ResNet34 | New dataset | Arabic |
| 9 | Modupe et al. (2022) | RDNN (CNN, BiLSTM) | CCAT50, IMDB62, Blogs50, Twitter50 | English |
| 10 | Jafariakinabad and Hua (2022) | Siamese LSTM | LAMBADA, CAT10/50, Blogs10/50 | English |
| 11 | Bauersfeld et al. (2023) | DistilBERT | New dataset | English |
| 12 | Hay et al. (2020) | DNN | Newspapers, Blogs | English |
| 13 | Khatun et al. (2020) | AWD-LSTM, mBERT, BanglaBERT | BAAD16, News, Wikipedia | Bangla |
| 14 | Najafi and Sadidpur (2024) | LSTM (attention) | New dataset | Persian |
| 15 | Theóphilo et al. (2019) | Char 4-grams, CNN | New dataset | English |
| 16 | Khatun et al. (2019) | Char-level CNN | New dataset | Bangla |
| 17 | Oldal and Kertész (2022) | CNN, GRU, LSTM | New dataset | Hungarian |
| 18 | Modupe et al. (2023) | BiLSTM-2D-CNN | CCAT50, IMDb62, Blog50, Twitter | English |
| 19 | Chowdhury et al. (2018b) | CNN, RNN (LSTM), MLP | New dataset | Bangla |
| 20 | Saha et al. (2018) | MLP | New dataset | English |
| 21 | Tang (2024) | CNN, LSTM | New dataset | Chinese |
| 22 | Bourib and Sayoud (2024) | CNN, RNN (LSTM, GRU) | C50 | English |
| 23 | Ranaldi et al. (2022) | BERT variants, FFNN | Islamic network forum | English |
| 24 | Etania and Riccosan (2023) | IndoBERT, M-BERT | Indonesian news | Indonesian |
| 25 | Demir and Can (2018) | R-ANN | New dataset | English |

AA using different variants of the BERT was done by AlZahrani and Al-Yahya (2023), Bauersfeld et al. (2023) and Etania and Riccosan (2023) for Arabic, English and Indonesian languages respectively. Bauersfeld et al. (2023) claim to have created the largest AA dataset from arXiv which can be used for AA based research.

They tested their DistilBERT based approach with different datasets and baselines and reported that while the DistilBERT-based method performs better than all baseline models on large datasets like Legal and IMDb62, it does not do better than basic n-gram models when working with small datasets. The explanation for these results was the data-hungry nature of transformers, which limits the performance of DistilBERT on smaller datasets. All deep learning-based research used embeddings for feature extraction and Suman et al. (2021), Bourib and Sayoud (2024), Modupe et al. (2023), Khatun et al. (2019) and Chowdhury et al. (2018b) tried to explore the impact of different embedding techniques in a quest to find the best for AA.

For word-level embeddings Suman et al. (2021) performed a series of experiments using different text representations, such as the GloVe embeddingss, BERT embeddings, character unigrams, and character bigrams. Their results showed that character unigrams and bigrams worked best with a capsule-based CNN model and even performed better than the current SOTA methods. Bourib and Sayoud (2024) also reported the superior performance of the Glove with LSTM and BiLSTM. Word level and character level embedding issues were discussed by Modupe et al. (2023), they experimented with sub word level embeddings and results indicated accuracy improvement of 1.07%, and 0.96%, on CCAT50 and Twitter, respectively. Khatun et al. (2019) however, proposed character level embeddings with CNN for Bangla and showed time and memory efficiency of their system over the word level counterpart with 2.5% less accuracy. Apart from embedding-based, feature extraction by lexical, stylometric and syntactic features with Deep Learning models was also investigated by some researchers. Silva and Frommholz (2023) combined average word length (AWL), average sentence length by word (ASW), and functional word counts (FWC) with BERT and performed a series of experiments by adding or removing one or a combination of these features with BERT. Results show that the best performance was achieved when BERT was used with all three features, demonstrating the importance of all three features. Superior performance of the models with added linguistic features was also reported by Saha et al. (2018), Najafi and Sadidpur (2024) and Ranaldi et al. (2022).

AA with machine-generated text as a class was also explored by Silva et al. (2024) and Silva and Frommholz (2023). In their experiments, Silva et al. (2024) used ChatGPT to forge novels written by human authors and then used it as an author class. The issues of closed world problems were discussed by Rahgouy et al. (2024), proposing CIL (class incremental learning) to address those issues where new authors are added gradually after the first training phase, which lets the system keep learning and adapting over time. The objective was to reorient AA researchers, the results are not promising but have the potential for adapting CIL in the field of AA.

Transfer learning (TL) can be used for domain adaptation, and it was explored by Khatun et al. (2020). They pre-trained on two different datasets, namely Wikipedia and news articles, and found that the Wikipedia dataset pre-training resulted in better downstream generalization.

All the experiments included in this section except one were based on supervised techniques. Jafariakinabad and Hua (2022) Introduced a self-supervised network containing two components: a lexical sub-network and a syntactic sub-network. Experiments were performed using both word and sentence embeddings and also by concatenating word embeddings with BoV, which increased the overall performance by 2.37% as compared to individual embedding settings. The summary of the articles with the datasets they used and the language they experimented with is given in Table 3.

3. **Traditional ML and DL**: Research involving both traditional ML- and DL-based approaches involves comparison of these approaches for specific settings, analysis of feature extraction techniques and classification techniques, combining different fea- tures for enhanced performance, combining classification models to achieve higher accuracy, sentiment analysis-based attribution, and use of TL for domain adaptation. The summary of this research is provided in table 4. Literature analysis shows that there are different feature extraction techniques like lexical, probabilistic, statistical syntactic, and semantic (Salau and Jain, 2019). Previous research was based on exploring these features and understanding their impact on the performance of the system. For the AA task, recent attempts involve experiments involving a number of different feature and classification models and finding the best combination of features and classification that can solve AA problem. Comparison of traditional ML and DL based models using a variety of different features was done by Alsanoosy et al. (2024), Fedotova et al. (2021), Misini, Canhasi, et al. (2024), Romanov et al. (2020), Misini, Kadriu, and Canhasi (2024), Al-Sarem et al. (2020) and Chowdhury et al. (2018a). Alsanoosy et al. (2024) compared six traditional ML models: Naïve Bayes (NB) , SVM, Decision Tree (DT), LR, K Nearest Neighbour (KNN), and Random Forest (RF) along with two DL architectures; Convolutional Neural Network (CNN) and Recurrent Neural Network (RNN). The best accuracy, 92.34%, was reached using an SVM model with TF-IDF features. The DL based CNN model had lower accuracy as compared to traditional ML models but still higher than the previously established baseline. Alsanoosy et al. (2024) used different feature extraction methods and manually selected the best based on accuracy. On the other hand, Fedotova et al. (2021) used genetic algorithms for feature selection. Their goal was optimization in terms of execution time, so they argued that although fastText accuracy is slightly lower, it is 51% faster than other DL based models. CNN based DL model have high accuracy but training time is higher than the combined time of the ML models. So in terms of good accuracy and less execution speed, the optimal variant is fastText. Similar studies like Chowdhury et al. (2018a) and Romanov et al. (2020), also reported the efficiency of fastText as it requires low computational power, model training and testing time.

Misini, Canhasi, et al. (2024) performed similar experiments for Albanian text and reported that the lexical features are the most effective set of linguistic fea- tures, significantly improving the performance of various algorithms in the AA task. According to Romanov et al. (2020) and Fedotova et al. (2021) the perfor- mance of ML models is low for short text and high for long text as compared to DL models. In an attempt towards AA in low-resource languages like Urdu, Nazir et al. (2021) claimed to have created the largest dataset called the Urdu News Authorship attribution Corpus (UNAAC-20). They tested it with various ML and DL models and reported that CNN is the most effective technique for the Urdu dataset. They stated that their dataset is much larger than existing ones and is also more difficult to use for the AA task.

**Table 4:** Summary of Research Papers on AA using ML and DL

| # | Article | Model | Dataset | Language |
|---|---|---|---|---|
| 1 | Z. Huang and Iwaihara (2022) | BERTweet, AWS, RoBERTA, ML models | Schwartz (Twitter) | English |
| 2 | Apoorva and Sangeetha (2021) | DNN, NB, SVM, K clustering | Enron Email | English |
| 3 | Wang and Iwaihara (2021) | RoBERTa, CNN | Twitter dataset | English |
| 4 | Chowdhury et al. (2018a) | FastText, NB, SVM | Blogs dataset | Bengali |
| 5 | Nitu and Dascalu (2024) | Romanian BERT, ML models | ROST extension | Romanian |
| 6 | Hossain et al. (2020) | MLP, SVM | New dataset | Bangla |
| 7 | Kori et al. (2022) | mBERT, KNN | New dataset | Multilingual |
| 8 | Briciu et al. (2021) | Autoencoder, ML classifiers | Romanian poems | Romanian |
| 9 | Alsanoosy et al. (2024) | NB, SVM, DT, RF, LR, KNN, CNN, RNN | Twitter dataset | English |
| 10 | Aouchiche et al. (2024) | TCN, LSTM+CNN, Autoencoder+Adaboost | Twitter dataset | English |
| 11 | Fedotova et al. (2021) | ML classifiers, LSTM, CNN, BERT, FastText | New dataset | Russian |
| 12 | Romanov et al. (2020) | SVM, LSTM, CNN | New dataset | Russian |
| 13 | Al-Sarem et al. (2020) | ML models, Feed-Forward NN | New dataset | Arabic |
| 14 | Misini, Kadriu, and Canhasi (2024) | XGBoost, ML classifiers, MLP | AAALit, CCAT10 | Albanian, English |
| 15 | Custódio and Paraboni (2021) | Ensemble: CNN, SVM, BERT | PAN18, lyrics, Twitter | Multilingual |
| 16 | Nazir et al. (2021) | ML models, LSTM, CNN | UrduCorpus, UNAAC-20 | Urdu |
| 17 | Avram and Oltean (2022) | ANN, ML classifiers | new dataset | Romanian |
| 18 | Misini, Canhasi, et al. (2024) | SVM, DT, CNN, XGB, FastText, mBERT | New dataset | Albanian |
| 19 | Hassan et al. (2017) | K-means clustering | PLOS dataset | English |
| 20 | Kocher and Savoy (2018) | KNN, chi-square, Delta, LDA, MLP, CBOW | Federalist, SOTU, Glasgow Herald | Multilingual |
| 21 | Siddiqui et al. (2023) | ML classifiers, CNN, GRU, LSTM, RoBERTa, BERT | New dataset | Urdu |
| 22 | Ivanov and Perez (2024) | MLP, SVM, RF, LMT | Gutenberg Poetry, Reuters-RCV1, English novels | English |

Combining the results of different individual classifiers can have a positive impact on the overall performance of the system. This idea was explored by Hossain et al. (2020) and Wang and Iwaihara (2021), where Hossain et al. (2020) used a voting system for individual results from MLP and SVM whereas Wang and Iwaihara (2021) used a hybrid approach, generating results from 2 modules, one by Roberta and CNN and the other style-based model using cosine similarity and triple loss objective functions. The output probabilities from the two modules are combined and then passed to a separate LR classifier for final classification. In both cases, an improvement in accuracy was reported. In similar experiments Custódio and Paraboni (2021) used stacked LR classifiers and Z. Huang and Iwaihara (2022) used combining classification models and different features based on UWS.

Most of the research involved supervised techniques, except for Apoorva and Sangeetha (2021) and Hassan et al. (2017) who used unsupervised techniques as well. One open problem that was reported by multiple research studies is decrease in performance when the number of authors are increased, but Apoorva and Sangeetha (2021) with their unsupervised clustering based approach demonstrated that the clustering technique is more effective in this specific scenario. Comparing results with supervised ML models, the clustering technique achieved 86% accu- racy on the entire dataset as compared to 76% using the supervised technique when using the entire dataset.

Features play an important role in classification as models learn on the basis of them. Kori et al. (2022) looked at both standalone embeddings (based on words, lemmas, PoS tags, and PoS masks) and composition-based embeddings, which were created by combining the standalone ones at the matrix level. They compared these methods and found that, for most of the languages studied, composition-based embeddings performed better than the baselines, including mBERT. Their main contribution was building a multilingual dataset of document representa- tions, which includes 28,204 documents (each with 10,000 tokens), 133 literary document embeddings across 7 European languages, and multi-language training weights organized by document type and language.

An interesting approach for attribution is to use sentiment analysis for attri- bution, which was done originally for a poetry corpus by Siddiqui et al. (2023) for Urdu Ghazals and Ivanov and Perez (2024) for English and produced promising results showing its potential for AA. Ivanov and Perez (2024) also tested it for non-poetry corpora like CEN (Corpus of English Novels) and a fiction-based cor- pus. The experiments with the CEN corpus indicated that sentiment analysis may indeed be useful for AA of non-poetry texts. While not as strong as the results from their poetry attribution experiments.

4. **LLMs based Research**: Leveraging LLMs for AA is relatively new with only 3 articles exploring LLMs for Attribution, highlighting the research gap. Table 5 shows the summary of the research using LLMs. AA using the GPT-2 model and perplexity score was done by W. Huang et al. (2024) and they argued about the superiority of perplexity over stylometry for the AA task. Their method outperformed SOTA for the BLOGS50 dataset but in the case of CCAT50 n-gram had better accuracy. Adewumi et al. (2024) worked on highlighting the limitations of LLMs in the AA task, given their tendency to hallucinate. They used LLaMA-2-13B, Mixtral 8x7B and Gemma-7B models to generate AA results. Mixtral showed superior performance and less hallucination, whereas Gemma had the lowest average accuracy. Barlas and Stamatatos (2021) explored the challenging yet realistic scenario of cross domain AA using transfer learning. They conducted a series of experiments for cross topic, cross genre and cross fandom AA. ELMo had high accuracy for cross-genre AA. BERT surpassed GPT 2 in all scenarios. BERT and ELMo achieved the highest accuracy.

**Table 5:** Summary of Research Papers on AA using LLMs

| # | Article | Model | Dataset | Language |
|---|---|---|---|---|
| 1 | Barlas and Stamatatos (2021) | BERT, GPT-2, ELMo, ULMFiT | CCMC, PAN18 | English |
| 2 | W. Huang et al. (2024) | GPT-2 | CCMC, PAN18 | English |
| 3 | Adewumi et al. (2024) | LlaMA, Mixtral, Gemma | New dataset | English |

### 3.1.1 Language Focus

Research in AA has been dominated by using the English language with 25 out of 59 articles involving solely English data, and 3 involving multilingual, including English. Efforts towards other languages and their AA have been made but they are very few. After English, Arabic and Bangla have the most articles, 4 each, Urdu and Romanian have 3 articles, and Indonesian and Russian have 2 articles each. Other languages which have been explored include Albanian, Portuguese, Sinhala, Latin Chinese, Hungarian and Persian with 1 article each.

Multilingual AA research has also been found and involves multiple language attribution using the same methods. The research by Kori et al. (2022) involves German, English, French, Hungarian, Portuguese, Slovenian, and Serbian. English and Albanian language attribution was explored by Misini, Kadriu, and Canhasi (2024). Custódio and Paraboni (2021) experimented on English, French, Italian, Polish, and Spanish. Efforts have been made towards attribution in other languages, but more effort is still needed. The absence of many languages in this section and the limited number of articles for non-English languages show a research gap in this direction.

### 3.1.2 Evolution of Methods for AA

Many researchers have studied the task of AA over the years. The use of DL models like CNN, BERT and LSTM have dominated the research field. Traditional ML models are still being used and produce promising results, especially for long texts. Exploring

LLMs for the task is relatively new but has the potential to be used more often in the future. With GenAI and new class for AA, leveraging LLMs to understand and differentiate between different authors, including machine-generated text is a new research direction. Research involving LLMs for AA started in 2024 and results show their potential in creating explainable and more accurate AA results.

## 3.2 Author Verification (AV)

AV is basically a binary classification task that aims to decide whether two documents were written by the same person or not (Boenninghoff et al., 2019). The task has evolved over time and has been used in different forms. The most prominent ones are:

1. Input 2 documents and verify if they are written by the same author.
2. Verify if the text is written by a human or machine (Guo et al., 2024).

There are significantly fewer AV-based research articles compared to AA. The methods and techniques employed for AV span from traditional ML models to more advanced encoder-decoder architectures and transformer-based models. With the rise of LLMs, their potential for this task has started to be explored. A brief overview of these is provided below.

1. **Traditional ML and DL**:

    Research in both traditional ML and DL often involves the use of multiple models simultaneously, reporting the best-performing model for specific settings. Studies that focused solely on traditional ML techniques include Castillo et al. (2019) using 2 variants of SVM classifier, namely C-SVC and Nu-SVC and Weeras- inghe et al. (2021) using LR. A notable aspect of Weerasinghe et al. (2021) is its application to open-set settings, where the authors included in the training set did not appear in the test set. Many studies experimented with multiple DL based methods. The most used model is BERT and its variants. Other important meth- ods used were CNN and RNNs. The summary of each article and the methods they used is provided in Table 6. Some researchers did experiments using both traditional ML and DL techniques and reported the results of which technique pro- duced the best results for their data or experimental setting. For instance, Khan et al. (2023) employed RF, SVM and CNN while Giorgi et al. (2020) used SVM, DT, RF, AdaBoost, SGD, RNN and CNN. Jiménez et al. (2023) tested LR, SVM and LSTM and finally Benzebouchi et al. (2018) used CNN, RNN and SVM. Both Khan et al. (2023) and Giorgi et al. (2020) found that the CNN outperformed tra- ditional ML methods. Benzebouchi et al. (2018) went beyond individual models by using majority voting of several classifiers, reporting improved performance when combining the results of all the classifiers. Giorgi et al. (2020) on the other hand, placed greater emphasis on feature extraction methods, finding that embedding- based feature extraction outperformed traditional feature engineering techniques in terms of accuracy.

2. **Large Language Model (LLM)**:

    The use of LLM for AV is relatively new, with only 4 out of 22 articles leveraging LLMs for this task. In X. Liu and Kong (2024), the authors used a strided sliding

**Table 6:** A summary of research on AV

| # | Article | Models Used | Dataset | Language |
|---|---|---|---|---|
| 1 | W. Huang and Grieve (2024) | GPT2, SVM | PAN 2024 | English |
| 2 | Guo et al. (2024) | BERT, BiLSTM | PAN | English |
| 3 | Futrzynski (2021) | BERT | CLEF 2020 | English |
| 4 | X. Liu and Kong (2024) | GPT2 | PAN | English |
| 5 | Brad et al. (2022) | BERT, Char BERT | PAN, DarkReddit | English |
| 6 | Weerasinghe et al. (2021) | LR | PAN (small, large) | English |
| 7 | Ramnath et al. (2024) | LLAMA3-8B (CAVE) | IMDb62, Blog-Auth, Fanfiction | English |
| 8 | Nguyen et al. (2023) | RoBERTa, ELECTRA, Longformer, BigBird | Fanfiction PAN 20/21 | English |
| 9 | J. Huang et al. (2024) | BERT | PAN | English |
| 10 | Ouyang et al. (2021) | Gated RNN | PAN 2015, Gutenberg, Enron, Reddit | English |
| 11 | Hu et al. (2023) | TDRLM | ICWSM, Twitter-Foursquare | English |
| 12 | Khan et al. (2023) | CNN, SVM, RF | UACV-22 | Urdu |
| 13 | R. Shao et al. (2024) | RoBERTa | New dataset | English |
| 14 | Giorgi et al. (2020) | RNN, CNN, LR, NN, SVM, DT, RF | Enron | English |
| 15 | Halvani et al. (2019) | 12 approches from PAN competition | New dataset | English |
| 16 | Khondaker et al. (2020) | BiLSTM | REUTER_C50, Spooky Author | English |
| 17 | Jiménez et al. (2023) | LR, SVM, LSTM | AuTexTification | English |
| 18 | Castillo et al. (2019) | SVM (C-SVC, Nu-SVC) | PAN 2014/15 | English |
| 19 | Jasper et al. (2018) | Deep LSTMs | PAN 2014/15 | English |
| 20 | Benzebouchi et al. (2018) | CNN, R-CNN, SVM | PAN 2015 | English |
| 21 | Hung et al. (2023) | GPT-3.5-Turbo | IMDB62 | English |
| 22 | Boenninghoff et al. (2019) | ADHOMINEM | New dataset | English |

window method with GPT-2 to get perplexity features. Their results showed that perplexity which measures how unpredictable a text is, can capture distinctive patterns characteristic of AI-generated content. The same feature perplexity was also utilized by W. Huang and Grieve (2024) combined with an SVM for final classification. It is believed that texts generated by LLMs usually have lower perplexity than human-written texts when measured using the same LLM. Both of these studies involved using GPT2 to measure perplexity score. In Hung et al. (2023) and Ramnath et al. (2024) the authors used prompt engineering techniques along with LLMs to generate the results. Hung et al. (2023) used GPT 3.5 turbo model and Ramnath et al. (2024) used offline LLAMA3-8B. Both of these approaches offered explainability, ensuring that the system's decisions could be trusted by users.

### 3.2.1 Language Focus

The overwhelming majority of the articles in the AV field focus on English, likely due to the availability of large, well-annotated datasets in English (e.g., PAN dataset, Enron email corpus). Additionally, English is a global lingua franca, so many of the challenges related to authorship and verification have been addressed in the context of English language data. For non-English AV, one article focused on AV in Urdu (Khan et al., 2023). No other articles with focus on any other language was found in literature which highlights a crucial research gap in the field.

### 3.2.2 Evolution of AV focus

In earlier studies, the focus was mainly on AV of human generated text, verifying if 2 documents were written by the same author. This was primarily driven by the need to analyze author styles across different works, detect plagiarism, and verify authorship in various contexts such as academic research, law, and literature. However, with the rise of generative AI tools and their widespread application in various forms of writing, from literary works to books and novels, the research focus has gradually shifted towards telling the difference between human-written and machine-generated texts. Notable studies addressing this emerging challenge include W. Huang and Grieve (2024), Guo et al. (2024), X. Liu and Kong (2024), J. Huang et al. (2024), Jiménez et al. (2023) and R. Shao et al. (2024).

## 3.3 Author attribution and verification

This section includes a total of 12 articles with focus on both sub-tasks, AA and AV. The detailed description of these articles is presented in Table 7.

1. **Traditional ML and DL**: The Ensembler approach using Deep Forest was employed by S. Shao et al. (2020) for internet relay chat (IRC) by monitoring online channels. Different channels like darkscience, computer, hak5 were monitored for data along with IRC public datasets provided by the AZSecure-data project. Profile-based and instance-based features were used for feature extraction. Studies involving DL include Saedi and Dras (2021) using a Siamese network (CNN as sub networks) and Fang et al. (2020) using BiLSTM.

   Evaluating both traditional ML and DL techniques together was done by Tyo et al. (2023), Reisi and Mahboob Farimani (2020), El-Halees (2022) and Romanov et al. (2024). Comparison of results generated by El-Halees (2022) using the BERT model and traditional ML models showed that BERT has superior performance for both AA and AV tasks. Moreover, the performance for sub-task AV is higher compared to AA. Similar results were reported by Reisi and Mahboob Farimani (2020) where they used a CNN and did comparison with traditional methods and found that CNN performed better. Romanov et al. (2024) experimented on AV of human and machine-generated text. The focus of Tyo et al. (2023) research was addressing inconsistent splits and mismatched evaluation methods. They created a benchmark called VALLA that standardizes AA/AV datasets and evaluation met- rics. It also includes a large-scale study and allows fair, direct comparisons between different methods. Another interesting study was conducted by Tripto et al. (2023) using fine-tuned BERT, GPT-who, and XGBoost. They created a paraphrased text dataset and compared the performance of current AA and AV models on the original text and paraphrased text and reported the diminishing performance in text classification models, with each successive paraphrasing iteration, provoking a reconsideration of the current notion of authorship.

**Table 7:** Summary of research involving both sub-tasks (AA and AV) using the same method

| # | Article | Method | Dataset | Language |
|---|---|---|---|---|
| 1 | B. Huang et al. (2024b) | Mistral 7B, Llama 2, GPT 3.5 turbo, GPT 4 turbo | Enron emails, Blog Authorship Corpus | English |
| 2 | Hicke and Mimno (2023) | LR, SVM (linear), cosine delta, Pythia, Falcon, fine-tuned T5 | Early English drama, EMED, SHC corpus | English |
| 3 | Tyo et al. (2023) | Ngram-based, BERT, PPM, pALM, HLSTM | CCAT50, CMCC, Guardian, IMDB62, blogs50, blogsall, PAN20/21, Amazon, Gutenberg, VALLA benchmark | English |
| 4 | Schmidt et al. (2024) | GPT-4o, Claude, Mistral-Large, Gemini-1.5 | Patristic Sermon Textual Archive (PaSTA) | Latin |
| 5 | Reisi and Mahboob Farimani (2020) | CNN, LSTM, SVM, K-means, Gaussian NB | Persian books | Persian |
| 6 | Tripto et al. (2023) | Fine-tuned BERT, GPT-who, XGBoost | New dataset | English |
| 7 | Saedi and Dras (2021) | Siamese network (CNN sub-network), BERT | PAN2015/18/20, IMDB, Enron emails | Multiple |
| 8 | S. Shao et al. (2020) | Ensemble method, Deep Forest | Online channels, AZSecure-data | English |
| 9 | Fang et al. (2020) | BiLSTM | Enron emails | English |
| 10 | El-Halees (2022) | Arabic-BERT, kNN, NB, SVM | Arabic poetry | Arabic |
| 11 | Romanov et al. (2024) | SVM, GRU, CNN, BERT, SimNN, QSUM | Student articles + GPT-4 texts | Russian |
| 12 | Huertas-Tato et al. (2024) | STAR, PART, RoBERTa | Blogs, Reddit TLDR, Twitter, Gutenberg | English |

2. **LLMs:** Schmidt et al. (2024) used LLMs for Latin AA and AV, focusing on historic language rather than modern language. They used LLMs such as GPT-4o, Gemini, Mistral, and Claude. GPT-4 performed best for AV so only that was used for the AA task. Other studies using LLMs include B. Huang et al. (2024b) using Mistral 7B, Llama 2, GPT 3.5 turbo and GPT 4 Turbo, and Hicke and Mimno (2023) using Pythia, Falcon and several fine-tuned T5. B. Huang et al. (2024b) designed LIP (linguistically informed prompt) and investigated how variations in the quantity and nature of instructions prompted to the LLMs impact the prediction's accuracy. They found that LLMs, equipped with the novel LIP technique, excel at identifying authorship without the need for domain-specific fine-tuning. Hicke and Mimno (2023) found T5 to be the best performing model. They also experimented with co-authored or disputed authored text and found that the model is greatly struggling to make clear attributions for such text.

### 3.3.1 Language Focus

English again was the prominent language for research in AV with 7 out of 12 articles focusing solely on it and 1 involving a multilingual study including English as one of the languages. Other languages involve Latin in Schmidt et al. (2024), Persian in Reisi and Mahboob Farimani (2020), Russian in Romanov et al. (2024) and Arabic in El-Halees (2022). Experiments on multiple languages at the same time were performed by Saedi and Dras (2021) who considered the English, Dutch, Spanish and Greek languages. The Pan 2015 AV dataset was used which is a multilingual dataset and experiments were performed using the same Siamese network on these languages with English having the highest accuracy. The key contribution was the use of a Siamese network for AA in datasets with large number of authors which is the limitation of most of the existing techniques.

## 4 Findings

This section highlights survey findings in terms of resources, hand crafted features, automatically extracted features, text type and languages found in literature and distribution of research over time. This will provide a better understanding of the the landscape to the new researchers who want to start research in this domain or topic.

### 4.1 Available Resources(data):

Data is arguably the most important component of any ML system and models are built on data. For authorship analysis every research study was based on data involving different languages, domains and sources (like blogs, social media, novels, books and articles). The benchmark datasets used for each sub-task are given below.

### 4.1.1 AA

AA datasets that have been widely used in the literature include;

1. **IMDB62:** A dataset comprising movie reviews from 62 prolific authors on IMDb, with over 1,000 reviews per author (Seroussi et al., 2014).
2. **PAN AA:** PAN is a series of scientific events and shared tasks focused on digital text forensics and writing style analysis (stylometry). It has many datasets of different sub-tasks. For AA it provides both close set (the author of disputed text belongs to known candidates) and open set (the author of disputed text may not necessarily belong to any known candidate) (Juola, 2012).
3. **Blog Authorship Corpus**: The Blog Authorship Corpus contains posts from 19,320 bloggers, collected from blogger.com in August 2004. It includes 681,288 posts with over 140 million words, averaging about 35 posts and 7,250 words per blogger (Schler et al., 2006).

4. **Reuters-50**: This dataset is a subset of the Reuters Corpus Volume 1 (RCV1) (Lewis et al., 2004), which has already been used in author identification studies. It includes the top 50 authors based on the total size of their articles, all of whom have written at least one article in the CCAT (corporate/industrial) category. The training set contains 2,500 texts (50 per author), and the test set includes another 2,500 non-overlapping texts (also 50 per author) (Z. Liu, 2006).
5. **CCAT50**: The CCAT50 dataset is a subset of the RCV1 (Lewis et al., 2004) and includes 5,000 financial news articles written by 50 authors, with each author having 50 training documents and 50 testing documents (Z. Liu, 2006).

### 4.1.2 AV

1. **PAN22 Authorship Analysis: AV**: This AV dataset is based on a new corpus in English which provides cross-Discourse type (DTs) verification cases. The DTs used are Essays, Emails, Text messages and Business memos. The focus of the dataset is on more challenging scenarios where each AV case considers two texts that belong to different DTs (cross-DT AV) (Stamatatos et al., 2022).
2. **Enron Email Dataset**: This dataset includes 500,000 emails written by employees of the Enron Corporation. It was collected by the Federal Energy Regulatory Commission during their investigation into Enron's collapse (Cohen, 2015).
3. **RCV1**: The RCV1 dataset is a well-known benchmark for text categorization. It includes news articles from Reuters written between 1996 and 1997. The dataset has 804,414 documents that were manually labeled and classified by industry, topic, and region using three controlled vocabularies (Lewis et al., 2004).
4. **AA Datasets**: In the literature, AA datasets such as The Blog authorship corpus, IMDB62 dataset have also been used for the AV task.

### 4.1.3 Other available datasets

Apart from the datasets mentioned above, new datasets have been developed in many research studies, for example, AA and AV for Low resource languages like Urdu, Bangla etc. and could be used for analysis. These datasets vary in language, size, text type, size of text, and instances per author. For low resource languages, where no datasets are available, contributions towards the creation of a balanced benchmark dataset is fundamental and can pave the way towards the future research. Datasets of different domains and text type for languages like English can also be beneficial to text methods for more challenging settings. Some of the datasets are given below.

1. **Indonesian News Article dataset**: This dataset was created by Etania and Riccosan (2023) and was also utilized by Saputra and Riccosan (2024). The dataset consists of articles obtained from Internet news portals, is compiled into a dataset in the Comma Separated Value (CSV) document format. It consists of 4,037 records of data, including 80 authors and one unknown.
2. **arXiv dataset**: This is a large dataset created from research papers available on arXiv, created by Bauersfeld et al. (2023) and contains over 2 million articles. Various trimmed versions of the dataset were also created for analysis. For example in one version they limited articles to 25 papers per author and called it D200T25-C. While it retains the 226 authors from the original D200-C dataset, the number of texts per author has been reduced from an average of 213 to 25. To analyze the impact of the number of authors on AA accuracy, a series of trimmed datasets was created from D200-C, featuring subsets with 200, 100, 50, and 25 texts per author.

3. **AAALitCorpus**: This dataset was created by Misini, Kadriu, and Canhasi (2024) and it consists of a collection of literary texts authored by prominent Albanian writers. The dataset comprises a total of 10 authors and 582 samples.
4. **UNAAC-20**: The Urdu News AA Corpus was developed by Nazir et al. (2021). It contains 26,118,475 words with a vocabulary of 291,728 unique tokens, and includes 21,938 articles written by 94 authors. The authors claim this is the largest AA dataset ever created for a low-resource language Urdu.
5. **UACV-22**: This dataset was built by gathering articles from Urdu newspapers to create a benchmark corpus for AV in the Urdu language. It was created by Khan et al. (2023) and was also utilized by Khan et al. (2024). The dataset comprises of 15 authors each with 400 articles.

The complete information of all these and other resources can be found in Tables 2-7. Comparison of the results and evaluation of the techniques using benchmark datasets provides better comparison as the same dataset is being used for each method. However, for experiments involving different data, or when a dataset is created, the analysis is not straight forward as there are many factors that contribute towards the performance of the model. Those factors include quality of the data, diversity, bias, size of text and number of authors and they can create bias in the results.

## 4.2 Feature Extraction

With DL the process of feature extraction is automatic as the embeddings automati- cally capture the syntactic and contextual meaning of the text. Some studies utilized the traditional methods as well. R. Shao et al. (2024) used character level N-gram and part-of-speech (PoS) tag N-gram, Weerasinghe et al. (2021) utilized frequency based approach TFIDF for extracting features and finally syntactic flow graph was used by Castillo et al. (2019) for feature extraction.

The feature extraction is important when using traditional ML methods as the model is built based on those. All DL models use embeddings for feature extraction, so we do not discuss each of them separately. However, a few studies did experi- ments with different feature extraction techniques, or different types of embedding and checked which one produced better results. Giorgi et al. (2020) experimented with embeddings and feature engineering and concluded that embedding based feature extraction produced better results. Khan et al. (2023) generated 9 embedding models for Urdu AV using a combination of three embedding techniques namely Word2Vec, GloVe and FastText. The significance of choosing the right embedding method and optimization algorithm for AV task in the Urdu literature was highlighted. The CNN- ADAM with fastText embedding performed best for the AV task with an accuracy of 98%. Suman et al. (2021) and Bourib and Sayoud (2024) explored different text representations under different settings and Bourib and Sayoud (2024) reported best performance using GloVe.

Different languages have different structures, hence the results could vary, a text representation performing well in one language might not work well with the other, and each method has some drawback. Issues with word and character level embeddings were discussed by Modupe et al. (2023). They proposed sub-word level embeddings and reported superior performance on different datasets. A few research studies combined many different features and also traditional feature extraction methods and dynamic DL based text representation methods and found that the combination has a positive impact on the performance. In one such experiment Nitu and Dascalu (2024) extracted contextualized representations from a pre-trained Romanian BERT model and concatenated these embeddings with linguistic features selected through the KruskalWallis mean rank, forming a hybrid input vector for a classification layer. Similarly Silva and Frommholz (2023) also combined hand-crafted stylometric features like Average Word Length (AWL), Average Sentence length by Word (ASW), and functional word counts (FWC) with BERT. The ablation study with only BERT and then adding and removing features, demonstrated that the best performance was achieved when using BERT plus all 3 features, demonstrating the importance of all 3 features. Other studies include Saha et al. (2018) combining linguistic features with Twitter features, Najafi and Sadidpur (2024) combining embeddings with syntactic features and Nitu and Dascalu (2024) who combined hand-crafted linguistic features such as word frequencies, syntax, semantics, and discourse markers with contextual embeddings from a Romanian BERT model. All these studies showed improvement in performance, as with more features the model can better understand the style and then classify the text.

## 4.3 Language and text type

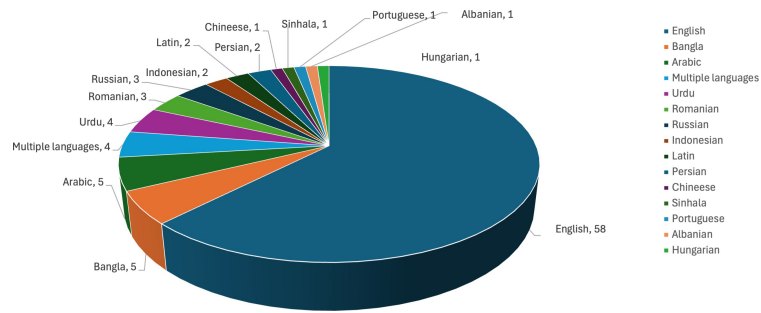

**Fig. 2:** Pie chart of the languages used in AA and AV based research

In the field of AA and AV, English-language research predominates, with numerous studies focusing on texts written in English and any other languages remain under-explored in this context. Tables 2 - 7 include information of each article and the language they used. Figure 2 illustrates the languages among the articles reviewed in descending order of the number of articles, showing that while English is the main focus. There is a significant lack of research on texts in other languages. This scarcity indicates a need for more comprehensive studies to achieve state-of-the-art results in AA and AV across diverse linguistic contexts. Expanding research to include a wider array of languages will enhance the robustness and applicability of attribution methods. In terms of text type, the datasets show a diverse range of text including both short texts from social media to large texts including research articles and books. Moreover the texts belong to different domains including fan fiction, literary text, research papers, PhD dissertations, Twitter posts, blogs, reviews, news articles, politics, religious data, stories, novels and official emails.

## 4.4 Evolution over time

This literature review focuses on articles published between 2015 and 2024. Table C3 in Appendix C presents the distribution of methodologies used across the surveyed literature. Due to inclusion and exclusion criteria mentioned in Section 2, no articles from 2015 and 2016 made it into the review as shown in Table C3 and Figure 3. Figure 3a shows the articles per year and what technique they used whereas 3b displays the distribution with respect to sub-task. From Figure 3a it is clear that from year 2017 to 2020, research predominantly utilized ML and DL techniques, with no reported use of LLMs. The use of LLMs is relatively new, where 2021 is the first instance of LLM application, accounting for 1 out of 14 articles. This trend continued, with LLM usage appearing in 2 out of 17 articles in 2023 and significantly rising to 7 out of 29 articles in 2024. This indicates a growing interest in integrating LLMs for AA and AV.

Prior to 2024, deep learning models dominated authorship analysis research. Interestingly even with the advent of LLMs and their powerful nature, the use of traditional ML and DL has dominated the research. Upon further analysis, it is evident that most studies leveraging LLMs employ pre-trained models in a zero-shot setting, often without fine-tuning and in context learning. While this has shown good results, it also suggests that the potential of LLMs in AA and AV is yet to be fully explored. Future research should explore fine-tuning, in context learning and domain adapta- tion techniques to enhance the performance and applicability of LLMs for these tasks. In terms of sub-task each year AA has dominated the research as compared to AV and it's shown in Figure 3b. Moreover the interest in using the same technique for both tasks together has also increased over the years. Finally, the total number of articles per year has also increased. This shows the growing interest in this research area, the open and challenging nature of the task and the new emerging challenges in this domain that arise with the advancement in technology.

## 4.5 Explainability

Explainability, also known as interpretability, refers to the ability to clearly understand and explain how a model arrives at its predictions or decisions in a human-comprehensible way (Linardatos et al., 2020). It is important if the models are to be used for solving problems where the user needs to trust the system like in the medical field, a legal system, courts, etc. Very few articles in the literature have touched this research direction where they used different techniques to explain the results. For the AA sub-task Silva and Frommholz (2023) explored it using LIME and Misini, Canhasi, et al. (2024) and Modupe et al. (2022) using SHapley Additive exPlanations (SHAP) analysis. Misini, Canhasi, et al. (2024) used SHAP values and waterfall plots to analyze how both handcrafted and model-generated features affect the author classification task. For AV Ramnath et al. (2024) introduced their model CAVE to generate structured and consistent explanations. The rationales are struc- tured as a set of sub-explanations grounded to AV-relevant linguistic features like sentence structure, punctuation style, etc. These sub-explanations have corresponding (intermediate) labels that provide further structure and also serve as a means to verify overall consistency with CAVEs prediction. The importance of explanation increases when models are adopted to be used in sensitive and critical situations like foren- sic analysis, the legal system and courts. Hence interpretability and explainability is important based on the domain and case study where we need to deploy the system.

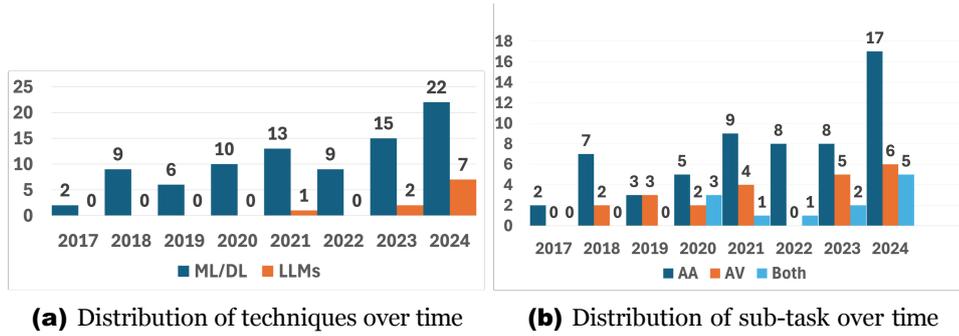

(a) Distribution of techniques over time  (b) Distribution of sub-task over time

**Fig. 3:** Distribution of articles over the years w.r.t (a) Techniques and (b)sub-task

### 4.6 Supervised vs Unsupervised Techniques

90 out of 93 articles utilized supervised learning (SL) for classification. Self supervised Learning (SSL) and unsupervised Learning (USL) were explored in 3 articles. SSL a subset of USL, where the model generates its own labels from the input data (Gui et al., 2024). USL is a technique where the model learns patterns and structures in the data without labeled outputs (Ziegel, 2003). In one of the studies Jafariakinabad and Hua (2022) used a self-supervised network containing two components: a lexical sub-network and a syntactic sub-network. Unsupervised methods were explored by Apoorva and Sangeetha (2021) and Hassan et al. (2017) both using K means cluster-ing. On a small number of authors the results were comparable, but when Apoorva and Sangeetha (2021) experimented with the entire Enron email dataset, USL achieved 86% accuracy compared to 76% using SL, demonstrating their potential to be used in open set setting or with increasing number of authors. This has the potential to be explored further especially when a large number of authors are involved in a dataset, as most of the studies reported a decrease in performance with an increase in the number of authors.

## 4.7 Multilingual Analysis

The literature review showed that the research in authorship analysis is dominated by the research in English-based text and very few other languages have been explored for this. Languages that have been explored include Bangla, Russian, Urdu and Latin, but with very limited research. Furthermore very few articles discussed the multilingual perspective, and the system's ability to adapt to different languages. Studies that involve multiple languages include Kori et al. (2022) for German, English, French, Hungarian, Portuguese, Slovenian, and Serbian, Misini, Kadriu, and Canhasi (2024) for Albanian and English, Custódio and Paraboni (2021) for English, French, Italian, Polish and Spanish. The adaptation of the model for different languages is not straight forward and it needs to consider the syntax, and morphological structure of the language. Multilingual research is identified as a critical research gap.

## 4.8 Exploring Machine-generated text

The most recent issue that has arisen is the use of generative AI for text generation, posing new challenges in this domain. This issue is addressed in both sub-tasks where in AA it is treated as an author class and for AV the task is to differentiate between human and machine-generated text. Very limited studies have addressed machine generated text. For AA Silva and Frommholz (2023) included machine-generated text, and differentiated between the two. The goal is multi-author attribution for a single document. The study showed that handcrafted stylometric features are effective. Using either the BERT model or stylometric features alone gave weaker results, but combining them led to a big improvement in performance.

For AV its a binary classification task where the goal is to differentiate between human and machine-generated text and it was studied by W. Huang and Grieve (2024), Guo et al. (2024), X. Liu and Kong (2024), J. Huang et al. (2024) and Jiménez et al. (2023). Traditional ML, DL and most advanced LLMs are used for machine generated text. In one of such studies, Jiménez et al. (2023), the authors proposed an approach that utilized traditional NLP feature extraction techniques focusing on linguistic properties, and traditional ML methods like LR and SSVM. They also compared the results with an ensemble of LSTM networks, each analyzing different paradigms of PoS tagging and claimed that their approach achieved a higher F1 score. DL-based machine generated text detection was done by Guo et al. (2024) and J. Huang et al. (2024). Guo et al. (2024) used BERT and BiLSTM, combining Transformer-based encoders and multiple text-feature extraction methods to enhance the model's ability to discriminate between texts. The proposed approach integrates pretrained BERT embeddings with linguistic features calculated by spaCy, further processed by BiLSTM and Transformer encoders for classification. On the other hand,
J. Huang et al. (2024) used a method called tri-sentence analysis (TSA), which helps the model better understand context and improves source identification. The text is split into short sections, each with three sentences, and each section is analyzed separately. These results are then combined to classify the whole text.

Finally LLM based studies include X. Liu and Kong (2024) and W. Huang and Grieve (2024), who used perplexity score for classification. X. Liu and Kong (2024) studied the differences between human-written and AI-generated texts, using a strided sliding window method with GPT-2 to extract perplexity features. W. Huang and Grieve (2024) introduced Authorial Language Models (ALMs) for AIAV, based on the idea that human-written texts usually have much higher perplexity than those created by LLMs. Use of machine generated text for both sub-tasks was studied by Tripto et al. (2023), where they discussed various experimental setups and iteration to answer an intriguing research question "Does a text retain its original authorship when it undergoes numerous paraphrasing iterations". They also talked about the ethical issues and raised some interesting questions about authorship after paraphrasing as to who should be the author after paraphrasing. The findings suggested that authorship should be task-dependent where in some tasks LLM paraphrasing should not alter authorship while in some others paraphrasing should change authorship. To conclude, this research direction has been explored a little but it still poses many challenges and is an open research area. With advancement and the use of LLMs for text generation, paraphrasing and content writing, AV is becoming an important task due to the widespread use of Gen AI and their potential to be used even more in the future.

## 4.9 Unfair Comparison

Comparing the results of different studies and declaring the superior performance of one technique over the other is not straight forward. A lot of factors are involved which need to be considered to evaluate the performance of a model. For example, text length, bias, diversity of dataset, number of instances per author, diversity of text per author and text type, all effect the performance of the model and comparison merely by performance metrics like accuracy shouldn't be considered absolute while comparing two methods. Results against benchmark datasets however can provide a better picture about the performance of the model. In the studies where new datasets are created, the general statement and claim of superior performance is debatable.

# 5 Future Directions

From the literature review the following directions have been identified where there is a need to explore more.

1. Research in AA and AV predominantly focuses on English, with limited studies addressing other languages. This imbalance highlights the need for future investigations to explore authorship analysis across diverse linguistic contexts. Expanding research to include a broader range of languages will enhance the robustness and applicability of attribution methods globally.
2. Leveraging LLMs for authorship analysis is new and has the potential to be explored more. Very few studies have utilized LLMs for authorship analysis. Using

LLMs for style analysis, fine tuning the model for the task and designing hybrid models could be an interesting research direction.
3. Very few researchers exploited USL and SSL. The K means clustering was used and claimed to have superior performance when the number of authors was increased. A problem where the majority of the literature had consensus is that the perfor- mance decreases with the increase in the number of authors. This could be further explored, to check its full potential for use on an increasing number of authors.
4. Explainability can play an important role in earning the trust of the users on the system. Few studies used explainability techniques along with the predictions like SHAP and Lime. This could be useful for tasks where understanding the results and trust is important, e.g in legal issues and court documents.
5. Machine generated text and it's identification is a relatively new and emerging sub-task. Its an open research question and potential research direction. With increasing use of LLMs and GenAI in many fields, more and more machine-generated text will be used in various domains. New methods and techniques need to be developed to address this emerging challenge.
6. The use of different datasets currently hinders the comparability and general- izability of different AA methods. Establishing comprehensive benchmarks that encompass a wide range of text types and sources, including human-authored, LLM-generated, and coauthored texts, would significantly enhance the field.
7. One of the key challenges in AA and AV is domain adaptation, where models trained on one type of text struggle to perform well on another. Since writing style can vary significantly across different domains, models that generalize across multiple domains are crucial for real-world applications. Domain adaptation is particularly important in cross-genre, cross-topic, and cross-linguistic settings, where traditional methods fail due to domain shifts in vocabulary, syntax, and discourse structures. A handful of studies explored this and more efforts are needed to create more adaptable, generalized systems for AA and AV.

# 6 Conclusion

This systematic literature review provides a comprehensive analysis of AA and AV, summarizing key methodologies, datasets, challenges, and emerging trends in the field. Over the years, ML and DL models have played a dominant role, while the recent integration of LLMs has introduced new possibilities for scalability and improved accuracy. While significant progress has been made, key challenges remain, particularly in multilingual adaptation, cross-domain generalization, explainability, and the detection of machine-generated text. This study underscores the importance of diverse and high-quality datasets, particularly for low-resource languages, and calls for further research in multilingual applications. By surveying existing research and identifying key gaps, this study provides a foundation for future advancements in authorship analysis. Con- tinued exploration of LLMs, hybrid feature extraction, explainable AI models, and adaptive learning techniques will be essential in developing scalable, fair, and more accurate AA and verification systems.

# Acknowledgments

This work is partly supported by the Wallenberg AI, Autonomous Systems and Soft- ware Program (WASP), funded by Knut and Alice Wallenberg Foundations and counterpart funding from Lulea University of Technology (LTU).

# Declarations

- Funding: Lulea University of Technology (LTU).
- Conflict of interest/Competing interests: No conflict of interests
- Ethics approval and consent to participate: Not applicable
- Consent for publication: Yes
- Data availability: Not applicable
- Materials availability: Provided in Appendix section
- Code availability: Not applicable
- Author contribution: The first author conducted the literature search and analy- sis drafting the manuscript, and updating the paper. The second contributed by reviewing the manuscript, providing feedback on structure and clarity, and offering academic guidance. Third author was involved in reviewing and refining the written work. The fourth supported the project in a supervisory capacity. She contributed to proofreading, drafting the layout of the paper and shaping the overall direction of the paper.
- Corresponding Author: Correspondence to Nudrat Habib

# Appendix A   Techniques used

The techniques used for AA and AV can be classified as

1. **Traditional ML based models:**

   - **Decision Tree (DT)**: A DT is a supervised learning algorithm utilized for clas- sification and regression tasks. It operates by recursively dividing the dataset based on feature values, creating a tree-like structure. In this structure, an internal node represents a decision on an attribute, branches indicate possible outcome, and each leaf node represents a class label (Quinlan, 1986).
   - **Naïve Bayes (NB)**: NB is a probabilistic classification algorithm that uses Bayes' theorem, assuming that all features are independent of each other given the class. This assumption makes the calculations easier and has proven useful in many tasks like text classification and spam detection (Kononenko, 1990).
   - **Logistic Regression (LR)**: LR is a statistical algorithm that fits a curve to the data using a logistic (sigmoid) function. The main goal of LR is to apply this nonlinear function to a set of linear features. It can be used for both binary and multiclass classification tasks (LaValley, 2008).
   - **Random Forest (RF)**: RF is an ensemble learning algorithm that creates many decision trees and combines their results to make better predictions and reduce overfitting (Breiman, 2001).

- **AdaBoost (Adaptive Boosting):** AdaBoost is an ensemble learning algorithm that improves the accuracy of weak classifiers by sequentially adjusting their weights. It assigns higher importance to misclassified instances in each iteration, allowing subsequent models to focus on difficult cases. AdaBoost is widely used for classification and regression tasks (Freund and Schapire, 1997).
- **K Nearest Neighbour (KNN)**: KNN is a simple, non-parametric, and lazy machine learning algorithm that can be used for both classification and regression tasks. It classifies a data point based on the majority class of its k nearest neighbors in feature space, making it simple yet effective for pattern recognition (Cover and Hart, 1967).
- **Support Vector Machine (SVM)**: SVM is a supervised learning algorithm introduced by Vapnik (1995) that is widely used for classification and regression tasks. SVM is particularly effective when the data is high-dimensional and has a clear margin of separation (Cortes and Vapnik, 1995).

2. **DL-based:** Besides ML, DL models have also been utilized for the task of AA and AV. The models that have been used and included in this review are:

   - **Convolutional Neural Network (CNN)**: A CNN is a special type of feed-forward neural network (also called a multi-layer perceptron or MLP) that is trained using backpropagation (Kuo, 2016).
   - **Recurrent Neural Network (RNN)**: RNNs are a class of neural networks that allow previous outputs to be used as inputs while having hidden states. They are widely used in tasks such as language modeling, speech recognition, and time-series forecasting but suffer from vanishing gradient issues in long sequences (Rumelhart et al., 1986).
   - **Bidirectional Encoder Representations from Transformers (BERT) and its variants**: Developed by Google, BERT is a pre-trained deep learning model introduced in 2018, designed to understand the context of words in a sentence by considering both left and right contexts (Devlin et al., 2019). Different variants of BERT used in literature are mBERT, DistillBERT, RoBERTa.
   - **Long short-term memory (LSTM) and its variants**: LSTM is a type of recurrent neural network (RNN) introduced by Hochreiter and Schmidhuber (1997). It was designed to manage long-term dependencies and reduce the van- ishing gradient problem, making it well-suited for tasks involving sequences, like speech recognition and language modeling Hochreiter and Schmidhuber (1997). Both LSTM and BiLSTM have been used in authorship analysis.

3. **LLM Based:**

   - **GPT-2**: GPT-2 is a transformer-based language model developed by OpenAI, containing 1.5 billion parameters. It is capable of generating coherent text and performing tasks such as translation and summarization without task-specific training (Radford et al., 2019).
   - **GPT-4**: OpenAIs GPT-4 is a multi-modal large language model that exhibits advanced reasoning, creativity, and contextual understanding, surpassing previous iterations in accuracy and coherence (OpenAI, 2023).

- **Mixtral**: Mixtral is a sparse Mixture of Experts (MoE) language model developed by Mistral AI, featuring 12 active experts per layer and significantly improving efficiency while maintaining high performance across various NLP tasks (Jiang et al., 2024).
- **Large Language Model Meta AI (LLaMA)**: Developed by Meta AI, it is a series of LLMs first introduced in February 2023. The latest iteration, Llama 3.2, released in September 2024, is a multi-modal model capable of processing both text and images, enhancing its versatility across various applications (Touvron et al., 2023).
- **Gemini**: Gemini, developed by Google DeepMind and launched on December 6, 2023, is a multi-modal large language model. It is the next step after Googles LaMDA and PaLM 2 models and can handle different types of data like text, images, audio, and video. Gemini powers Googles AI applications, such as the Bard chatbot and AI Mode in Google Search, positioning it as a competitor to models like OpenAIs GPT-4 (Team et al., 2023).
- **Claude:** Claude is a series of advanced AI language models developed by Anthropic, designed for natural language understanding, reasoning, and ethi- cal AI interactions. Using Constitutional AI, it aligns responses with predefined ethical principles to ensure helpfulness and safety.
- **Gemma**: Developed by Google DeepMind, Gemma is an open-weight large language model introduced in February 2024, featuring 2B and 7B parameter variants optimized for diverse computational environments (Team et al., 2024).
- **Falcon**: Developed by the Technology Innovation Institute (TII), Falcon is a series of open-weight LLMs, with Falcon 180B (2023) being one of the largest, featuring 180 billion parameters and trained on 3.5 trillion tokens (Penedo et al., 2023).
- **Pythia**: Developed by EleutherAI, is a collection of 16 autoregressive language models with sizes ranging from 70 million to 12 billion parameters. (Biderman et al., 2023).

4. **Unsupervised techniques**: Very few articles used unsupervised techniques for classification purposes. The method used was K-means clustering.

   - **K-means clustering** is an unsupervised learning algorithm that divides a dataset into K groups by minimizing the distance between points and their group centers. The algorithm works by repeatedly assigning points to the closest clus- ter center (centroid) and then updating the centroids based on the average of the points in each group (MacQueen, 1967).

5. Combination of the above techniques.

# Appendix B  Evidence of methodology

**Table B1:** Primary Search Terms Used in the Systematic Review

| Keyword ID | Search Term |
|---|---|
| K1 | Large Language Models |
| K2 | Machine Learning |
| K3 | Deep Learning |
| K4 | Author Attribution |
| K5 | Author Verification |
| K6 | Text |

**Table B2:** Search Strings Used in Each Database and Corresponding Links

| Database | Search String | Link (if available) |
|---|---|---|
| Scopus | K1 and K4 AND K6 | https://www.scopus.com/results/results.uri?sort=plf-f&src=s&sid=4d7bfb2d6d7b0c477298f7571c64eee0&sot=b&sdt=b&sl=142&s=%28TITLE-ABS-KEY%28Large+language+models%29+AND+TITLE-ABS-KEY%28author+AND+attribution%29+AND+TITLE-ABS-KEY%28text%29%29+AND+PUBYEAR+%3E+2014+AND+PUBYEAR+%3C+2025&origin=savedSearchNewOnly&txGid=05e56f4353b1e07a4032c3ff3eee283c&sessionSearchId=4d7bfb2d6d7b0c477298f7571c64eee0&limit=10 |
| | K1 and K5 AND K6 | https://www.scopus.com/results/results.uri?sort=plff&src=s&sid=91661a4786c962b3c5ef73947de0ad12&sot=b&sdt=b&sl=147&s=%28TITLE-ABS-KEY%28large+AND+language+AND+models%29+AND+TITLE-ABS-KEY%28author+verification%29+AND+TITLE-ABS-KEY%28text%29%29+AND+PUBYEAR+%3E+2014+AND+PUBYEAR+%3C+2025&origin=savedSearchNewOnly&txGid=6ae222edca6dc8c19bccf427b7d61b02&sessionSearchId=91661a4786c962b3c5ef73947de0ad12&limit=10 |
| | (K2 or K3) and K4 AND K6 | https://www.scopus.com/results/results.uri?sort=plf-f&src=s&sid=143475b4a69fe4d6bc8b74dca5d38c00&sot=b&sdt=b&sl=173&s=%28TITLE-ABS-KEY%28machine+AND+learning%29+OR+TITLE-ABS-KEY%28deep+AND+learning%29+AND+TITLE-ABS-KEY%28author+attribution%29+AND+TITLE-ABS-KEY%28Text%29%29+AND+PUBYEAR+%3C+2025&origin=savedSearchNewOnly&txGid=4069e3cc731d8041dda2e27a4105214e&sessionSearchId=143475b4a69fe4d6bc8b74dca5d38c00&limit=10 |
| | (K2 or K3) and K5 AND K6 | https://www.scopus.com/results/results.uri?sort=plf-f&src=s&sid=c1a8758f73eb1ecb984241dfc8bebf18&sot=b&sdt=b&sl=178&s=%28TITLE-ABS-KEY%28machine+AND+learning%29+OR+TITLE-ABS-KEY%28deep+AND+learning%29+AND+TITLE-ABS-KEY%28author+AND+verification%29+AND+TITLE-ABS-KEY%28text%29%29+AND+PUBYEAR+%3E+2014+AND+PUBYEAR+%3C+2025&origin=savedSearchNewOnly&txGid=e659627aa26b2e7e767ee3273c8aac11&sessionSearchId=c1a8758f73eb1ecb984241dfc8bebf18&limit=10 |
| | K1 AND K4 OR K5 AND K6 | https://www.scopus.com/results/results.uri?sort=plf-f&src=s&sid=140018f254880344c5b1b6532e623e93&sot=b&sdt=b&sl=179&s=%28TITLE-ABS-KEY%28Large+language+model%29+AND+TITLE-ABS-KEY%28author+attribution%29+OR+TITLE-ABS-KEY%28author+AND+verification%29+AND+TITLE-ABS-KEY%28text%29%29+AND+PUBYEAR+%3E+2014+AND+PUBYEAR+%3C+2025&origin=savedSearchNewOnly&txGid=5e483b230366be1b00eb3dc7520941d7&sessionSearchId=140018f254880344c5b1b6532e623e93&limit=10 |
| | K3 AND K4 OR K5 AND K6 | https://www.scopus.com/results/results.uri?sort=plf-f&src=s&sid=fefe764ad532c97d8cdacfb350937812&sot=b&sdt=b&sl=176&s=%28TITLE-ABS-KEY%28Deep+learning%29+AND+TITLE-ABS-KEY%28author+attribution%29+OR+TITLE-ABS-KEY%28author+AND+verification%29+AND+TITLE-ABS-KEY%28text%29%29+AND+PUBYEAR+%3E+2014+AND+PUBYEAR+%3C+2025&origin=savedSearchNewOnly&txGid=6d23475cddf8930fed59630bddb58172&sessionSearchId=fefe764ad532c97d8cdacfb350937812&limit=10 |
| Google Scholar | K1 AND K4 AND K6 | https://scholar.google.com/scholar?hl=en&as_sdt=0%2C5&as_ylo=2015&as_yhi=2024&q=%22Large+language+models%22+AND+%22author+attribution%22+AND+%22Text%22&btnG= |
| | K1 AND K5 AND K6 | https://scholar.google.com/scholar?hl=en&as_sdt=0%2C5&as_ylo=2015&as_yhi=2024&q=%22Large+language+models%22+AND+%22author+verification%22+AND+%22Text%22&btnG= |
| | (K2 or K3) AND K4 AND K6 | https://scholar.google.com/scholar?hl=en&as_sdt=0%2C5&as_ylo=2015&as_yhi=2024&q=%22Machine+Learning%22+OR+%22Deep+Learning%22+AND+%22author+attribution%22+AND+%22Text%22&btnG= |
| | (K2 or K3) AND K5 AND K6 | https://scholar.google.com/scholar?hl=en&as_sdt=0%2C5&as_ylo=2015&as_yhi=2024&q=%22Machine+Learning%22+OR+%22Deep+Learning%22+AND+%22author+verification%22+AND+%22Text%22&btnG= |
| | K1 AND K4 OR K5 AND K6 | https://scholar.google.com/scholar?hl=en&as_sdt=0%2C5&as_ylo=2015&as_yhi=2024&q=%22Large+language+models%22+AND+%22Author+attribution%22+OR+%22author+verification%22+AND+%22Text%22&btnG= |
| | K3 AND K4 OR K5 AND K6 | https://scholar.google.com/scholar?hl=en&as_sdt=0%2C5&as_ylo=2015&as_yhi=2024&q=%22Deep+Learning%22+AND+%22author+attribution%22+OR+%22author+verification%22+AND+%22Text%22&btnG= |

# Appendix C    Results

From the Table C3, the values in sub-tasks are consistent with total number of articles but for technique it exceeds 93. This is due to the overlap of techniques between LLM and ML/DL as shown in figure 1 where 4 articles are overlapping. Hence total of 97 here instead of 93.

**Table C3:** Distribution over the years

| Year | No of articles | Technique | | Sub-tasks | | |
|---|---|---|---|---|---|---|
| | | ML / DL | LLM | AA | AV | Both |
| 2017 | 2 | 2 | 0 | 2 | 0 | 0 |
| 2018 | 9 | 9 | 0 | 7 | 2 | 0 |
| 2019 | 6 | 6 | 0 | 3 | 3 | 0 |
| 2020 | 10 | 10 | 0 | 5 | 2 | 3 |
| 2021 | 13 | 14 | 1 | 9 | 4 | 1 |
| 2022 | 9 | 9 | 0 | 8 | 0 | 1 |
| 2023 | 15 | 15 | 2 | 8 | 5 | 2 |
| 2024 | 28 | 22 | 7 | 17 | 6 | 5 |
| | | | | | | |
| Total | 93 | 87 | 10 | 59 | 22 | 12 |

# References

Adewumi, T., Habib, N., Alkhaled, L., & Barney, E. (2024). On the limitations of large language models (LLMs): False attribution. *arXiv preprint arXiv:2404.04631*.

Alsanoosy, T., Shalbi, B., & Noor, A. (2024). Authorship attribution for English short texts. *Engineering, Technology & Applied Science Research*, *14* (5), 16419–16426.

Al-Sarem, M., Alsaeedi, A., & Saeed, F. (2020). A deep learning-based artificial neural network method for instance-based Arabic language authorship attribution. *Int J Adv Soft Comput Appl*, *12* (2), 1–15.

Al-Sarem, M., Saeed, F., Qasem, S. N., & Albarrak, A. M. (2023). Deep learning-based method for enhancing the detection of Arabic authorship attribution using acoustic and textual-based features. *International Journal of Advanced Computer Science and Applications*, *14* (7).

Altakrori, M. H., Iqbal, F., Fung, B. C., Ding, S. H., & Tubaishat, A. (2018). Arabic authorship attribution: An extensive study on twitter posts. *ACM Transactions on Asian and Low-Resource Language Information Processing (TALLIP)*, *18* (1), 1–51.

AlZahrani, F. M., & Al-Yahya, M. (2023). A transformer-based approach to authorship attribution in classical Arabic texts. *Applied Sciences*, *13* (12), 7255.

Aouchiche, R. I. A., Boumahdi, F., Remmide, M. A., & Madani, A. (2024). Authorship attribution in twitter: A comparative study of machine learning and deep learning approaches. *International Journal of Information Technology*, *16* (5), 3303–3310.

Apoorva, K., & Sangeetha, S. (2021). Deep neural network and model-based clustering technique for forensic electronic mail author attribution. *SN Applied Sciences*, *3* (3), 348.

Avram, S.-M., & Oltean, M. (2022). A comparison of several AI techniques for authorship attribution on Romanian texts. *Mathematics*, *10* (23), 4589.

Barlas, G., & Stamatatos, E. (2021). A transfer learning approach to cross-domain authorship attribution. *Evolving Systems*, *12* (3), 625–643.

Bauersfeld, L., Romero, A., Muglikar, M., & Scaramuzza, D. (2023). Cracking double-blind review: Authorship attribution with deep learning. *Plos One*, *18* (6), e0287611.

Benzebouchi, N. E., Azizi, N., Aldwairi, M., & Farah, N. (2018). Multi-classifier system for authorship verification task using word embeddings. *2018 2nd International Conference on Natural Language and Speech Processing (ICNLSP)*, 1–6.

Biderman, S., Schoelkopf, H., Anthony, Q. G., Bradley, H., OBrien, K., Hallahan, E., Khan, M. A., Purohit, S., Prashanth, U. S., Raff, E., et al. (2023). Pythia: A suite for analyzing large language models across training and scaling. *International Conference on Machine Learning*, 2397–2430.

Boenninghoff, B., Hessler, S., Kolossa, D., & Nickel, R. M. (2019). Explainable authorship verification in social media via attention-based similarity learning. *2019 IEEE International Conference on Big Data (Big Data)*, 36–45.

Bourib, S., & Sayoud, H. (2024). Author identification using LSTM and BiLSTM techniques applied on imbalanced data set with hyper encoding parameters. *2024 IEEE International Conference on Advanced Systems and Emergent Technologies (IC_ASET)*, 1–6.

Brad, F., Manolache, A., Burceanu, E., Barbalau, A., Ionescu, R. T., & Popescu, M. (2022). Rethinking the authorship verification experimental setups. *Proceedings of the 2022 Conference on Empirical Methods in Natural Language Processing*, 5634–5643.

Breiman, L. (2001). Random forests. *Machine learning*, *45*, 5–32.

Briciu, A., Czibula, G., & Lupea, M. (2021). Autoat: A deep autoencoder-based classification model for supervised authorship attribution. *Procedia Computer Science*, *192*, 397–406.

Castillo, E., Cervantes, O., & Vilarino, D. (2019). Authorship verification using a graph knowledge discovery approach. *Journal of Intelligent & Fuzzy Systems*, *36* (6), 6075–6087.

Chowdhury, H. A., Imon, M. A. H., & Islam, M. S. (2018a). Authorship attribution in Bengali literature using FastText's hierarchical classifier. *2018 4th International Conference on Electrical Engineering and Information & Communication Technology (iCEEiCT)*, 102–106.

Chowdhury, H. A., Imon, M. A. H., & Islam, M. S. (2018b). A comparative analy- sis of word embedding representations in authorship attribution of Bengali literature. *2018 21st International Conference of Computer and Information Technology (ICCIT)*, 1–6.

Cohen, W. W. (2015). Enron email dataset. https://www.loc.gov/item/2018487913/

Cortes, C., & Vapnik, V. (1995). Support-vector networks. *Machine learning*, *20*, 273–297.

Cover, T., & Hart, P. (1967). Nearest neighbor pattern classification. *IEEE transactions on information theory*, *13* (1), 21–27.

Custódio, J. E., & Paraboni, I. (2021). Stacked authorship attribution of digital texts. *Expert Systems with Applications*, *176*, 114866.

da Rocha, M. A., de Morais, P. S. G., da Silva Barros, D. M., dos Santos, J. P. Q., Dias-Trindade, S., & de Medeiros Valentim, R. A. (2022). A text as unique as a fingerprint: Text analysis and authorship recognition in a Virtual Learning Environment of the Unified Health System in Brazil. *Expert Systems with Applications*, *203*, 117280.

Demir, N. M., & Can, M. (2018). Authorship authentication of short messages from social networks machines. *Southeast Europe Journal of Soft Computing*, *7*(1).

Devlin, J., Chang, M.-W., Lee, K., & Toutanova, K. (2019). Bert: Pre-training of deep bidirectional transformers for language understanding. *Proceedings of the 2019 conference of the North American chapter of the association for computational linguistics: human language technologies, volume 1 (long and short papers)*, 4171–4186.

El, S. E. M., & Kassou, I. (2014). Authorship analysis studies: A survey. *International Journal of Computer Applications*, *86* (12).

El-Halees, A. M. (2022). Arabic poetry authorship attribution and verification using transfer learning. *Egyptian Computer Science Journal*, *46* (1).
Ellen, D., Day, S., & Davies, C. (2018). *Scientific examination of documents: Methods and techniques*. CRC Press.
Etania, S. K., & Riccosan, C. A. (2023). Automatic Indonesian authorship attribution recognition using transformer. , *17* (05), 497.
Fang, Y., Yang, Y., & Huang, C. (2020). Emaildetective: An email authorship identification and verification model. *The Computer Journal*, *63* (11), 1775–1787.
Fedotova, A., Romanov, A., Kurtukova, A., & Shelupanov, A. (2021). Authorship attribution of social media and literary Russian-language texts using machine learning methods and feature selection. *Future Internet*, *14* (1), 4.
Fourkioti, O., Symeonidis, S., & Arampatzis, A. (2019). Language models and fusion for authorship attribution. *Information Processing & Management*, *56* (6), 102061.
Freund, Y., & Schapire, R. E. (1997). A decision-theoretic generalization of on-line learning and an application to boosting. *Journal of computer and system sciences*, *55* (1), 119–139.
Futrzynski, R. (2021). Author classification as pre-training for pairwise authorship verification. *CLEF (Working Notes)*, 1945–1952.
Giorgi, G., Saracino, A., & Martinelli, F. (2020). Email spoofing attack detection through an end to end authorship attribution system. *ICISSP*, 64–74.
Gui, J., Chen, T., Zhang, J., Cao, Q., Sun, Z., Luo, H., & Tao, D. (2024). A survey on self-supervised learning: Algorithms, applications, and future trends. *IEEE Transactions on Pattern Analysis and Machine Intelligence*, *46* (12), 9052–9071. https://doi.org/10.1109/TPAMI.2024.3415112
Guo, L., Yang, W., Ma, L., & Ruan, J. (2024). BLGAV: generative AI author verification model based on BERT and BiLSTM. *Working Notes of CLEF*.
Halvani, O., Winter, C., & Graner, L. (2019). Assessing the applicability of author- ship verification methods. *Proceedings of the 14th International Conference on Availability, Reliability and Security*, 1–10.
Halvani, O., Winter, C., & Pflug, A. (2016). Authorship verification for different languages, genres and topics. *Digital Investigation*, *16*, S33–S43.
Hassan, S.-U., Imran, M., Iftikhar, T., Safder, I., & Shabbir, M. (2017). Deep stylometry and lexical & syntactic features based author attribution on PLoS digital repository. *International conference on Asian digital libraries*, 119–127.
Hay, J., Doan, B.-L., Popineau, F., & Elhara, O. A. (2020). Filtering a reference corpus to generalize stylometric representations. *KDIR*, 259–268.
He, X., Lashkari, A. H., Vombatkere, N., & Sharma, D. P. (2024). Authorship attribution methods, challenges, and future research directions: A comprehensive survey. *Information*, *15* (3), 131.
Hicke, R. M., & Mimno, D. (2023). T5 meets Tybalt: Author Attribution in Early Modern English Drama Using Large Language Models. *arXiv preprint arXiv:2310.18454*.

Hochreiter, S., & Schmidhuber, J. (1997). Long short-term memory. *Neural computation*, *9*(8), 1735–1780.

Hossain, A. S., Akter, N., & Islam, M. S. (2020). A stylometric approach for author attribution system using neural network and machine learning classifiers. *Proceedings of the International Conference on Computing Advancements*, 1–7.

Hu, X., Ou, W., Acharya, S., Ding, S. H., DGama, R., & Yu, H. (2023). TDRLM: Stylometric learning for authorship verification by Topic-Debiasing. *Expert Systems with Applications*, *233*, 120745.

Huang, B., Chen, C., & Shu, K. (2024a). Authorship attribution in the era of LLMs: Problems, methodologies, and challenges. *arXiv preprint arXiv:2408.08946*.

Huang, B., Chen, C., & Shu, K. (2024b). Can large language models identify authorship? *arXiv preprint arXiv:2403.08213*.

Huang, J., Chen, Y., Luo, M., & Li, Y. (2024). Generative AI authorship verification of tri-sentence analysis base on the BERT model. *Working Notes of CLEF*.

Huang, W., & Grieve, J. (2024). Authorial language models for AI authorship verification. *Working Notes of CLEF*.

Huang, W., Murakami, A., & Grieve, J. (2024). ALMs: Authorial Language Models for Authorship Attribution. *arXiv preprint arXiv:2401.12005*.

Huang, Z., & Iwaihara, M. (2022). Capsule network over pre-trained language model and user writing styles for authorship attribution on short texts. *Proceedings of the 2022 3rd International Conference on Control, Robotics and Intelligent System*, 104–110.

Huertas-Tato, J., Martín, A., & Camacho, D. (2024). Understanding writing style in social media with a supervised contrastively pre-trained transformer. *Knowledge-Based Systems*, *296*, 111867.

Hung, C.-Y., Hu, Z., Hu, Y., & Lee, R. K.-W. (2023). Who wrote it and why? prompting large-language models for authorship verification. *arXiv preprint arXiv:2310.08123*.

Ivanov, L., & Perez, F. (2024). Authorship Attribution of English Poetry using Sentiment Analysis. *The International FLAIRS Conference Proceedings*, *37*.

Jafariakinabad, F., & Hua, K. A. (2022). A self-supervised representation learning of sentence structure for authorship attribution. *ACM Transactions on Knowledge Discovery from Data (TKDD)*, *16*(4), 1–16.

Jalal Yousef Zaidieh, A. (2024). Combatting cybersecurity threats on social media: Network protection and data integrity strategies. *Journal of Artificial Intelligence and Computational Technology*, *1*(1).

Jasper, J., Berger, P., Hennig, P., & Meinel, C. (2018). Authorship verification on short text samples using stylometric embeddings. *Analysis of Images, Social Networks and Texts: 7th International Conference, AIST 2018, Moscow, Russia, July 5–7, 2018, Revised Selected Papers 7*, 64–75.

Jiang, A. Q., Sablayrolles, A., Roux, A., Mensch, A., Savary, B., Bamford, C., Chaplot, D. S., de las Casas, D., Hanna, E. B., Bressand, F., Lengyel, G., Bour,

G., Lample, G., Lavaud, L. R., Saulnier, L., Lachaux, M.-A., Stock, P., Subramanian, S., Yang, S., . . . Sayed, W. E. (2024). Mixtral of experts. https://arxiv.org/abs/2401.04088

Jiménez, D., Cardoso-Moreno, M. A., Aguilar-Canto, F., Juárez-Gambino, O., & Calvo, H. (2023). Poslemma: How traditional machine learning and linguistics preprocessing aid in machine generated text detection. *Computación y Sistemas*, *27*(4), 921–928.

Juola, P. (2012). An overview of the traditional authorship attribution subtask. *CLEF (Online Working Notes/Labs/Workshop)*, *1178*, 1.

Khan, T. F., Anwar, W., Arshad, H., & Abbas, S. N. (2023). An Empirical Study on Authorship Verification for Low Resource Language using Hyper-Tuned CNN Approach. *IEEE Access*.

Khan, T. F., Sabir, M., Malik, M. H., Ghous, H., Ijaz, H. M., Nadeem, A., & Ejaz, A. (2024). Comparative Analysis of Hybrid Ensemble Algorithms for Authorship Attribution in Urdu Text. *Journal of Computing & Biomedical Informatics*.

Khatun, A., Rahman, A., Islam, M. S., et al. (2019). Authorship Attribution in Bangla literature using Character-level CNN. *2019 22nd International Conference on Computer and Information Technology (ICCIT)*, 1–5.

Khatun, A., Rahman, A., Islam, M. S., Chowdhury, H. A., & Tasnim, A. (2020). Authorship attribution in Bangla literature (AABL) via transfer learning using ULMFiT. *Transactions on Asian and Low-Resource Language Information Processing*.

Khondaker, M. T. I., Khan, J. Y., Alam, T., & Rahman, M. S. (2020). Agree-to-disagree (a2d): A deep learning-based framework for authorship discrimination task in corpus-specificity free manner. *IEEE Access*, *8*, 162322–162334.

Kocher, M., & Savoy, J. (2018). Distributed language representation for authorship attribution. *Digital Scholarship in the Humanities*, *33*(2), 425–441.

Kononenko, I. (1990). Comparison of inductive and naive bayesian learning approaches to automatic knowledge acquisition. *Current trends in knowledge acquisition*, *8*, 190.

Kuo, C.-C. J. (2016). Understanding convolutional neural networks with a mathematical model. *Journal of Visual Communication and Image Representation*, *41*, 406–413.

Lastowka, G. (2007). Digital attribution: Copyright and the right to credit. *BUL Rev.*, *87*, 41.

LaValley, M. P. (2008). Logistic regression. *Circulation*, *117*(18), 2395–2399.

Lewis, D. D., Yang, Y., Rose, T. G., & Li, F. (2004). Rcv1: A new benchmark collection for text categorization research. *Journal of machine learning research*, *5*(Apr), 361–397.

Linardatos, P., Papastefanopoulos, V., & Kotsiantis, S. (2020). Explainable ai: A review of machine learning interpretability methods. *Entropy*, *23*(1), 18.

Liu, X., & Kong, L. (2024). AI text detection method based on perplexity features with strided sliding window. *Working notes of clef*.

Liu, Z. (2006). Reuter$_5$0$_5$0 [DOI: https://doi.org/10.24432/C5DS42].

MacQueen, J. (1967). Some methods for classification and analysis of multivariate observations. *Proceedings of the Fifth Berkeley Symposium on Mathematical Statistics and Probability, Volume 1: Statistics*, *5*, 281–298.

Marchenko, O., Anisimov, A., Nykonenko, A., Rossada, T., & Melnikov, E. (2017). Authorship attribution system. In F. Frasincar, A. Ittoo, L. M. Nguyen, & E. Métais (Eds.), *Natural language processing and information systems* (pp. 227–231). Springer International Publishing.

Misini, A., Canhasi, E., Kadriu, A., & Fetahi, E. (2024). Automatic authorship attribution in Albanian texts. *Plos One*, *19*(10), e0310057.

Misini, A., Kadriu, A., & Canhasi, E. (2024). Authorship classification techniques: Bridging textual domains and languages. *International Journal on Information Technologies & Security*, *16*(1).

Modupe, A., Celik, T., Marivate, V., & Olugbara, O. O. (2022). Post-authorship attribution using regularized deep neural network. *Applied Sciences*, *12*(15), 7518.

Modupe, A., Celik, T., Marivate, V., & Olugbara, O. O. (2023). Integrating bidirectional long short-term memory with subword embedding for authorship attribution. *2023 IEEE International Conference on Systems, Man, and Cybernetics (SMC)*, 1910–1917.

Najafi, M., & Sadidpur, S. (2024). Paa: Persian author attribution using dense and recursive connection. *Preprints*.

Nazir, Z., Shahzad, K., Malik, M. K., Anwar, W., Bajwa, I. S., & Mehmood, K. (2021). Authorship attribution for a resource poor languageUrdu. *Transactions on Asian and Low-Resource Language Information Processing*, *21*(3), 1–23.

Nguyen, T., Dagli, C., Alperin, K., Vandam, C., & Singer, E. (2023). Improving long-text authorship verification via model selection and data tuning. *Proceedings of the 7th Joint SIGHUM Workshop on Computational Linguistics for Cultural Heritage, Social Sciences, Humanities and Literature*, 28–37.

Nitu, M., & Dascalu, M. (2024). Authorship Attribution in Less-Resourced Languages: A Hybrid Transformer Approach for Romanian. *Applied Sciences*, *14*(7), 2700.

Oldal, L. G., & Kertész, G. (2022). Evaluation of Deep Learning-based Authorship Attribution Methods on Hungarian Texts. *2022 IEEE 10th Jubilee International Conference on Computational Cybernetics and Cyber-Medical Systems (ICCC)*, 000161–000166.

OpenAI. (2023). GPT-4 technical report. *arXiv preprint arXiv:2303.08774*. https://arxiv.org/abs/2303.08774

Ouni, S., Fkih, F., & Omri, M. N. (2023). A survey of machine learning-based author profiling from texts analysis in social networks. *Multimedia Tools and Applications*, *82*(24), 36653–36686.

Ouyang, L., Zhang, Y., Liu, H., Chen, Y., & Wang, Y. (2021). Gated POS-level language model for authorship verification. *Proceedings of the Twenty-Ninth International Conference on International Joint Conferences on Artificial Intelligence*, 4025–4031.

Penedo, G., Malartic, Q., Hesslow, D., Cojocaru, R., Cappelli, A., Alobeidli, H., Pannier, B., Almazrouei, E., & Launay, J. (2023). The RefinedWeb dataset for Falcon LLM: Outperforming curated corpora with web data, and web data only. *arXiv preprint arXiv:2306.01116*. https://arxiv.org/abs/2306.01116

Peng, J., Choo, K.-K. R., & Ashman, H. (2016). Bit-level n-gram based forensic authorship analysis on social media: Identifying individuals from linguistic profiles. *Journal of Network and Computer Applications*, *70*, 171–182.

Quinlan, J. R. (1986). Induction of decision trees. *Machine learning*, *1*, 81–106.

Raafat, M. A., El-Wakil, R. A.-F., & Atia, A. (2021). Comparative study for stylometric analysis techniques for authorship attribution. *2021 International Mobile, Intelligent, and Ubiquitous Computing Conference (MIUCC)*, 176–181.

Radford, A., Wu, J., Child, R., Luan, D., Amodei, D., Sutskever, I., et al. (2019). Language models are unsupervised multitask learners. *OpenAI blog*, *1* (8), 9.

Rahgouy, M., Giglou, H. B., Tabassum, M., Feng, D., Das, A., Rahgooy, T., Dozier, G., & Seals, C. D. (2024). Towards effective authorship attribution: Integrating class-incremental learning. *2024 IEEE 6th International Conference on Cognitive Machine Intelligence (CogMI)*, 56–65.

Ramnath, S., Pandey, K., Boschee, E., & Ren, X. (2024). Cave: Controllable authorship verification explanations. *arXiv preprint arXiv:2406.16672*.

Ranaldi, L., Ranaldi, F., Fallucchi, F., & Zanzotto, F. M. (2022). Shedding light on the dark web: Authorship attribution in radical forums. *Information*, *13* (9), 435.

Reisi, E., & Mahboob Farimani, H. (2020). Authorship attribution in historical and literary texts by a deep learning classifier. *Journal of Applied Intelligent Systems and Information Sciences*, *1* (2), 118–127.

Ribeiro, R., Carvalho, J. P., & Coheur, L. (2024). Leveraging fuzzy fingerprints from large language models for authorship attribution. *2024 IEEE International Conference on Fuzzy Systems (FUZZ-IEEE)*, 1–7.

Ríos-Toledo, G., Velázquez-Lozada, E., Posadas-Duran, J. P. F., Prado Becerra, S., Pech May, F., & Monjaras Velasco, M. G. (2023). Evaluation of feature extraction techniques in automatic authorship attribution. *Computación y Sistemas*, *27* (2), 371–377.

Romanov, A., Kurtukova, A., Shelupanov, A., Fedotova, A., & Goncharov, V. (2020). Authorship identification of a Russian-language text using support vector machine and deep neural networks. *Future Internet*, *13* (1), 3.

Romanov, A., Shelupanov, A., Kurtukova, A., & Fedotova, A. (2024). Integrated technique of natural and artificial texts authorship verification in the academic environment.

Rosen-Zvi, M., Chemudugunta, C., Griffiths, T., Smyth, P., & Steyvers, M. (2010). Learning author-topic models from text corpora. *ACM Transactions on Information Systems (TOIS)*, *28* (1), 1–38.

Rumelhart, D. E., Hinton, G. E., & Williams, R. J. (1986). Learning representations by back-propagating errors. *nature*, *323* (6088), 533–536.

Saedi, C., & Dras, M. (2021). Siamese networks for large-scale author identification. *Computer Speech & Language*, *70*, 101241.

Saha, N., Das, P., & Saha, H. N. (2018). Authorship attribution of short texts using multi-layer perceptron. *International Journal of Applied Pattern Recognition*, *5* (3), 251–259.

Salau, A. O., & Jain, S. (2019). Feature extraction: A survey of the types, tech- niques, applications. *2019 international conference on signal processing and communication (ICSC)*, 158–164.

Saputra, K., & Riccosan, R. (2024). Indonesian news article authorship attribution multilabel multiclass classification using indobert. *IAES International Jour- nal of Artificial Intelligence (IJ-AI)*, *13* (4), 4688–4694. https://doi.org/10.11591/ijai.v13.i4.pp4688-4694

Sarwar, R., Perera, M., Teh, P. S., Nawaz, R., & Hassan, M. U. (2024). Crossing linguistic barriers: authorship attribution in Sinhala texts. *ACM Transactions on Asian and Low-Resource Language Information Processing*, *23* (5), 1–14.

Schler, J., Koppel, M., Argamon, S., & Pennebaker, J. W. (2006). Effects of age and gender on blogging. *AAAI spring symposium: Computational approaches to analyzing weblogs*, *6*, 199–205.

Schmidt, G., Gorovaia, S., & Yamshchikov, I. P. (2024). Sui Generis: Large Language Models for Authorship Attribution and Verification in Latin. *arXiv preprint arXiv:2410.09245*.

Seroussi, Y., Zukerman, I., & Bohnert, F. (2014). Authorship attribution with topic models. *Computational Linguistics*, *40* (2), 269–310.

Shao, R., Schwarz, R., Clifton, C., & Delp, E. (2024). A natural approach for synthetic short-form text analysis. *Proceedings of the 2024 Joint International Con- ference on Computational Linguistics, Language Resources and Evaluation (LREC-COLING 2024)*, 1034–1042.

Shao, S., Tunc, C., Al-Shawi, A., & Hariri, S. (2020). An ensemble of ensem- bles approach to author attribution for internet relay chat forensics. *ACM Transactions on Management Information Systems (TMIS)*, *11* (4), 1–25.

Siddiqui, I., Rubab, F., Siddiqui, H., & Samad, A. (2023). Poet Attribution of Urdu Ghazals using Deep Learning. *2023 3rd International Conference on Artificial Intelligence (ICAI)*, 196–203.

Silva, K., & Frommholz, I. (2023). What if chatgpt wrote the abstract?-explainable multi-authorship attribution with a data augmentation strategy. *IACT@ SIGIR*, 38–48.

Silva, K., Frommholz, I., Can, B., Blain, F., Sarwar, R., & Ugolini, L. (2024). Forged- GAN-BERT: Authorship Attribution for LLM-Generated Forged Novels. *Proceedings of the 18th Conference of the European Chapter of the Association for Computational Linguistics: Student Research Workshop*, 325–337.

Kori, M., Stankovi, R., Ikoni Nei, M., Byszuk, J., & Eder, M. (2022). Parallel stylo- metric document embeddings with deep learning based language models in literary authorship attribution. *Mathematics*, *10* (5), 838.

Stamatatos, E., Kredens, K., Pezik, P., Heini, A., Kestemont, M., Bevendorff, J., Pot- thast, M., & Stein, B. (2022, March). *Pan22 authorship analysis: Authorship verification* (Version 0.1.1). Zenodo. https://doi.org/10.5281/zenodo.6337151

Suman, C., Raj, A., Saha, S., & Bhattacharyya, P. (2021). Authorship attribution of microtext using capsule networks. *IEEE Transactions on Computational Social Systems*, *9*(4), 1038–1047.

Tang, X. (2024). Author identification of literary works based on text analysis and deep learning. *Heliyon*, *10*(3).

Team, G., Anil, R., Borgeaud, S., Alayrac, J.-B., Yu, J., Soricut, R., Schalkwyk, J., Dai, A. M., Hauth, A., Millican, K., et al. (2023). Gemini: A family of highly capable multimodal models. *arXiv preprint arXiv:2312.11805*.

Team, G., Mesnard, T., Hardin, C., Dadashi, R., Bhupatiraju, S., Pathak, S., Sifre, L., Rivière, M., Kale, M. S., Love, J., et al. (2024). Gemma: Open models based on gemini research and technology. *arXiv preprint arXiv:2403.08295*.

Theóphilo, A., Pereira, L. A., & Rocha, A. (2019). A needle in a haystack? harnessing onomatopoeia and user-specific stylometrics for authorship attribution of micro-messages. *ICASSP 2019-2019 IEEE International Conference on Acoustics, Speech and Signal Processing (ICASSP)*, 2692–2696.

Touvron, H., Lavril, T., Izacard, G., Martinet, X., Lachaux, M.-A., Lacroix, T., Rozière, B., Goyal, N., Hambro, E., Azhar, F., et al. (2023). Llama: Open and efficient foundation language models. *arXiv preprint arXiv:2302.13971*.

Tripto, N. I., Venkatraman, S., Macko, D., Moro, R., Srba, I., Uchendu, A., Le, T., & Lee, D. (2023). A Ship of Theseus: Curious Cases of Paraphrasing in LLM-Generated Texts. *arXiv preprint arXiv:2311.08374*.

Tyo, J., Dhingra, B., & Lipton, Z. C. (2022). On the state of the art in authorship attribution and authorship verification. *arXiv preprint arXiv:2209.06869*.

Tyo, J., Dhingra, B., & Lipton, Z. C. (2023). Valla: Standardizing and benchmarking authorship attribution and verification through empirical evaluation and comparative analysis. *Proceedings of the 13th International Joint Conference on Natural Language Processing and the 3rd Conference of the Asia-Pacific Chapter of the Association for Computational Linguistics (Volume 1: Long Papers)*, 649–660.

Wang, X., & Iwaihara, M. (2021). Integrating roberta fine-tuning and user writing styles for authorship attribution of short texts. *Web and Big Data: 5th International Joint Conference, APWeb-WAIM 2021, Guangzhou, China, August 23–25, 2021, Proceedings, Part I 5*, 413–421.

Watson, G. (1966). Frederick Mosteller, David L. Wallace, Inference and Disputed Authorship: The Federalist. *The Annals of Mathematical Statistics*, *37*(1), 308–312.

Weerasinghe, J., Singh, R., & Greenstadt, R. (2021). Feature vector difference based authorship verification for open-world settings. *CLEF (Working Notes)*, 2201–2207.

Zhang, O. R., Cohen, T., & McGill, S. (2018). Did Gaius Julius Caesar Write De Bello Hispaniensi? A Computational Study of Latin Classics Authorship. *Human IT: Journal for Information Technology Studies as a Human Science*, *14*(1), 28–58.

Ziegel, E. R. (2003). The elements of statistical learning.